\documentclass{article}

 \usepackage[main, final]{neurips_2026}

\usepackage[utf8]{inputenc} 
\usepackage[T1]{fontenc}    
\usepackage{hyperref}       
\usepackage{url}            
\usepackage{booktabs}       
\usepackage{amsfonts}       
\usepackage{nicefrac}       
\usepackage{microtype}      
\usepackage{xcolor}         

\usepackage{amsmath}
\usepackage{amssymb}
\usepackage{mathtools}
\usepackage{amsthm}
\usepackage{microtype}
\usepackage{graphicx}
\usepackage{subcaption}
\usepackage{booktabs} 

\usepackage[all]{nowidow}

\usepackage{booktabs}
\usepackage{tabularx}
\usepackage{ragged2e}
\newcolumntype{Y}{>{\RaggedRight\arraybackslash}X}
\usepackage{amssymb} 

\usepackage{booktabs}
\usepackage{multirow}
\usepackage{makecell}
\usepackage{siunitx}

\title{EnvDreamer: Large-Scale Multimodal-to-Environment Generation for Embodied AI}

\author{%
  Kabir Swain \\
  MIT \\
  \texttt{kswain@mit.edu} \\
  \And
  Sijie Han \\
  University of Toronto \\
  \texttt{hs.han@mail.utoronto.ca} \\
  \And
  Antonio Torralba \\
  MIT \\
  \texttt{torralba@mit.edu} \\
}

\begin{document}

\maketitle

\begin{abstract}
Large datasets and high capacity models have accelerated progress in vision and language. This work introduces a platform aimed at bringing comparable gains to embodied learning, world models, and robotics. We present EnvDreamer, a framework that uses large language and vision language models to generate Unreal Engine 5 environments for embodied AI and robot training. EnvDreamer enables sampling of large, diverse, interactive, customizable, and validator passed virtual environments for training and evaluation across navigation, interaction, and manipulation. We illustrate the platform with a large set of generated scenes and simple baselines. Policies trained on EnvDreamer generated environments, without explicit mapping or human task supervision, achieve competitive results on multiple embodied benchmarks spanning navigation, rearrangement, and manipulation. EnvDreamer also supports image-conditioned reconstruction for real-to-sim studies. Finally, we release EnvDreamer-20k, a dataset of 20,000 validator passed environments with task programs, scene graphs, trajectories, and metadata to support reproducible benchmarking.
\end{abstract}
\section{Introduction}

\begin{figure*}[h]
    \centering
    \includegraphics[width=\textwidth]{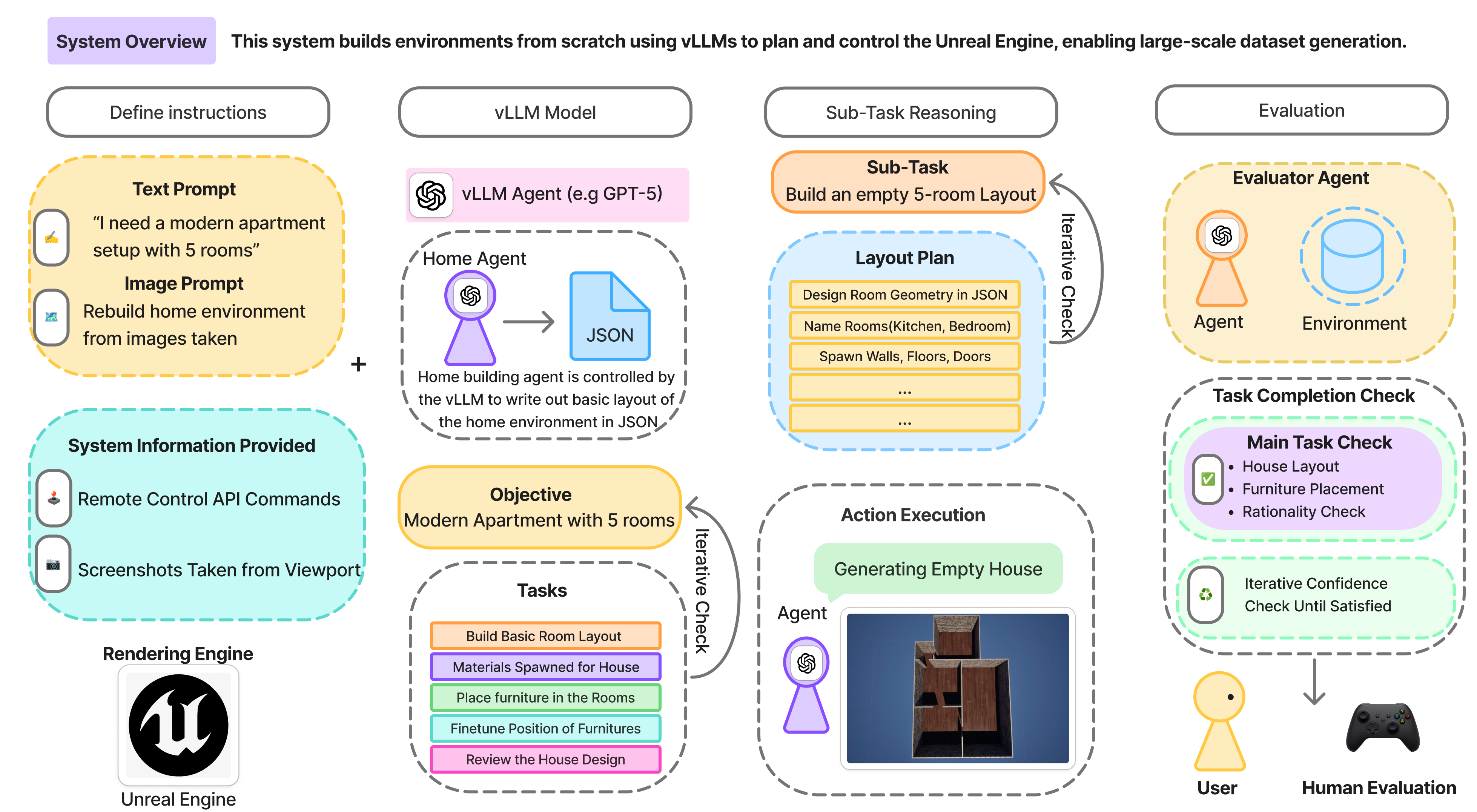}
    \caption{\textbf{EnvDreamer system overview.} The system builds environments from scratch using vision–language models (vLLMs) to plan and control Unreal Engine~5. User-provided text or image prompts along with system information such as Remote Control API commands and viewport screenshots (left) are passed to a vLLM-based Home Agent which decomposes the request into sub-tasks with Chain of Thought\cite{wei2022chainofthought}, generates a structured layout plan and task list in JSON, and compiles them into EnvDreamer API Python code for execution (middle). An Evaluator Agent then performs iterative task-completion checks on scene layout, furniture placement, and rationality, providing feedback until the environment passes validation (right). Finally, users can inspect the completeness of the environment by entering a play session in either an ego-centric or bird’s-eye view.}
    \label{fig:system_overview}
\end{figure*}

Large scale training on diverse data has accelerated progress in vision and language. In contrast, embodied AI and world model research still rely on simulators with far fewer scenes and limited task diversity, which makes it difficult to study long horizon behavior and robust generalization. Although recent simulators have improved physics, manipulation, and sensor realism, scaling interactive environments to tens or hundreds of thousands of scenes while preserving quality and reproducibility remains challenging~\cite{igibson2021}~\cite{manipulathor2021}. Pipelines based on manual authoring or 3D scans are expensive to expand by orders of magnitude and are not designed for language first task authoring or rapid iteration~\cite{matterport3d2017} ~\cite{scannet2017}.

We present \textbf{EnvDreamer} (Figure ~\ref{fig:system_overview}), a framework that generates fully interactive, physics enabled Unreal Engine 5 environments for embodied AI, world models, and robotics. Given a natural language description and optionally images, EnvDreamer plans assets, layout, and task specifications; compiles these plans to Unreal Python; and iteratively revises scenes and task logic using simulator feedback with unit style validators that check playability, physics consistency, and goal completeness. A large asset library with semantic categories, materials, and articulated parts is used to populate scenes automatically while enforcing physical plausibility and commonsense constraints. Lighting, textures, and materials can vary on load to increase diversity and reduce overfitting. Beyond text to scene synthesis~\cite{text2scene2019}~\cite{housegan2020}~\cite{rplan2019}~\cite{sceneformer2021}, EnvDreamer supports image conditioned generation for sim to real studies, allowing researchers to reconstruct target environments in simulation for training and evaluation.

To demonstrate scalability, we construct \textbf{EnvDreamer 20k}, a collection of 20{,}000 validator passed environments that span indoor and outdoor settings and include executable task programs, scene graphs, trajectories, and metadata. We train simple baselines for world model pretraining on interactive rollouts and for embodied policy learning in navigation, rearrangement, and manipulation using RGB and other modest sensory inputs, without explicit mapping or human task supervision. We report competitive results across multiple embodied benchmarks and study transfer by pretraining on EnvDreamer and evaluating in zero shot and few shot settings. For robotics, we aim to train policies that interact with objects to complete tasks in EnvDreamer and hope to deploy them to real hardware; we also plan to examine sim to real performance when simulators are fit to target scenes via image conditioned reconstruction.

In summary, our contributions are: (1) \textbf{EnvDreamer}, a language first, code generative framework for diverse, fully interactive UE5 environments with unit style validators for playability, physics consistency, and goal completeness; (2) \textbf{EnvDreamer 20k}, a public dataset of 20{,}000 validator passed environments with executable task programs, scene graphs, trajectories, and metadata for reproducible benchmarking across indoor and outdoor settings; (3) \textbf{Comprehensive evaluation} showing competitive results for world model pretraining and embodied policy learning with analyses of zero shot and few shot transfer relative to procedural and non interactive baselines.


\section{Related Work}

\paragraph{Embodied AI platforms.}
A range of simulators support embodied research with different emphases and tradeoffs. VirtualHome~\cite{DBLP:journals/corr/abs-1806-07011} focuses on programmatic household activities with scene graphs and scripted routines. AI2\textendash THOR~\cite{DBLP:journals/corr/abs-1712-05474} offers interactive objects, object states, and accurate physics across common indoor rooms, and ProcTHOR~\cite{deitke2022procthor} extends it with large scale procedural generation of apartments and houses. Habitat~\cite{DBLP:journals/corr/abs-1904-01201} prioritizes fast simulation and 3D scanned scenes, historically with less emphasis on richly editable object dynamics. OmniGibson/BEHAVIOR~\cite{li2024behavior1khumancenteredembodiedai} advances physics grounded interaction and task definitions. Across these platforms, scene creation is typically manual or template procedural, and task specifications are often goal descriptions rather than executable programs with explicit preconditions and success predicates. EnvDreamer differs by using language as the primary specification, compiling to Unreal Python, and applying validator checks for playability, physics consistency, and goal completeness before release.

\paragraph{LLM evaluation in embodied environments.}
Closely related is VirtualEnv~\cite{virtualenv2026}, an Unreal Engine~5 platform for evaluating LLM-driven embodied agents on navigation, manipulation, and procedurally generated tasks such as escape rooms. EnvDreamer builds on this platform, extending it with language-driven environment generation, unit-style validators for playability and physics consistency, and a large-scale dataset designed for embodied policy and world-model pretraining.

\begin{table*}[ht]
\centering
\small
\renewcommand{\arraystretch}{1.2}
\setlength{\tabcolsep}{0pt}
\begin{tabular*}{\textwidth}{@{\extracolsep{\fill}}lcccccc@{}}
\toprule
\textbf{System} & \textbf{Interactive} & \textbf{Language} & \textbf{Real-to-Sim} & \textbf{Task Type} & \textbf{Scale (Envs)} & \textbf{Open-Vocab} \\
\midrule
ProcTHOR~\cite{deitke2022procthor}        & \checkmark &            &            & CS   & 10k+               &            \\
Infinigen Indoors~\cite{raistrick2024infinigen} &            &            &            & S    & 5k+                &            \\
PGDrive~\cite{li2021pgdrive}              & \checkmark &            &            & CST  & 100k+              &            \\
Phone2Proc~\cite{deitke2023phone2proc}    & \checkmark &            & \checkmark & SH   & 100+/scan          &            \\
Holodeck~\cite{yang2024holodeck}          & \checkmark & \checkmark &            & CSH  & 1k+                & \checkmark \\
Architect~\cite{zhang2024architect}       &            & \checkmark &            & CS   & \(\sim\)100/category & \checkmark \\
\textbf{EnvDreamer (Ours)}                & \checkmark & \checkmark & \checkmark & \textbf{CSTH} & \textbf{20k} & \checkmark \\
\bottomrule
\end{tabular*}
\caption{\textbf{Comparison of procedural environment generation systems.}
\emph{Interactive}: agent embodiment with physics and actuation. 
\emph{Language}: prompt-guided scene generation. 
\emph{Real-to-Sim}: conditioning on real scenes to rebuild them in simulation 
\emph{Task Type}: coverage across constraint-free generation (C), spatial/layout diversity (S), temporal dynamics (T), and heterogeneous domains (H). 
\emph{Scale}: approximate number of distinct environments. 
\emph{Open-Vocab}: support for unbounded prompt categories at environment generation. 
EnvDreamer achieves full CSTH coverage with scalable, language-driven generation and UE5-based interactivity.}
\label{tab:envdreamer_cvprstyle_final}
\end{table*}

\begin{figure*}[t]
    \centering
    \includegraphics[width=\textwidth]{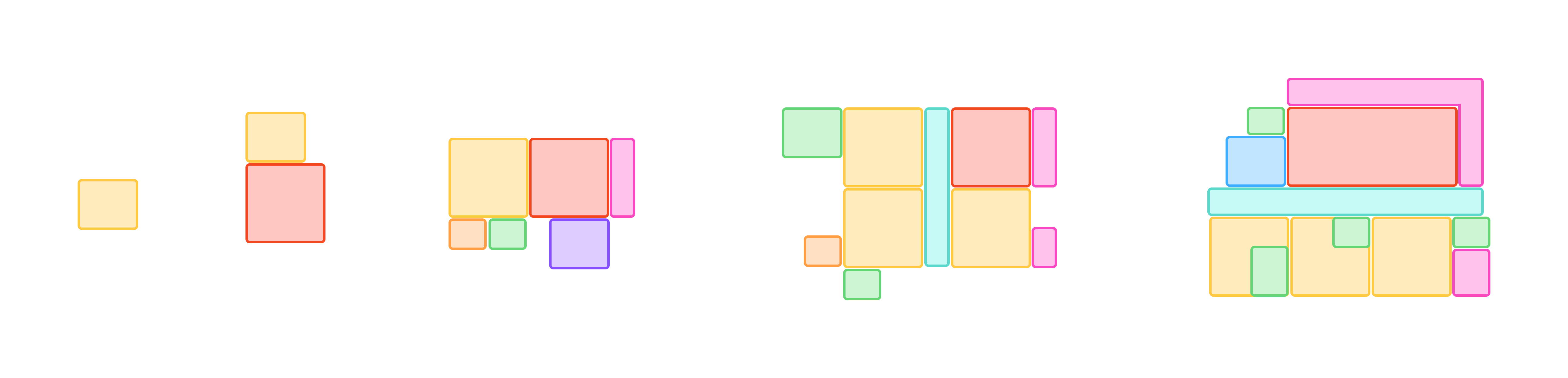}
    \caption{\textbf{Diverse and realistic floor plan generation with EnvDreamer.} 
    Unlike traditional procedural methods that rely on fixed generation pipelines with limited variation, EnvDreamer leverages language-driven synthesis to create diverse, realistic home layouts. Each layout shown was generated from natural language prompts (e.g., "Create a modern 2-bedroom apartment," "Design a spacious family home with open kitchen"), producing architectural arrangements that follows real-world constraints. The diversity spans from compact studios (left) to complex multi-room residences (right), with rooms color-coded by type: bedrooms (yellow), bathrooms (green), living areas (red), and other spaces. This prompt-based approach enables researchers to specify exact requirements for embodied AI and robotics experiments while maintaining architectural realism that fixed procedural pipelines struggle to achieve.}
    \label{fig:floorplan_diversity}
\end{figure*}

\paragraph{Large scale datasets.}
Broad datasets have propelled progress in vision and language; the embodied setting has fewer interactive corpora of comparable scale. ProcTHOR~\cite{deitke2022procthor} procedurally generates indoor houses to enable arbitrarily many scenes. Scanned repositories used with Habitat offer visual realism but are expensive to expand and less suited to program level task authoring. BEHAVIOR and related OmniGibson efforts~\cite{li2024behavior1khumancenteredembodiedai} target household skills with physics and object states. EnvDreamer complements these directions by providing language driven scene and task synthesis, unit style validation at generation time, and multimodal creation, and by releasing \textbf{EnvDreamer 20k}, a dataset of 20{,}000 validator passed environments with task programs, scene graphs, trajectories, and metadata for reproducible benchmarking across indoor and outdoor settings.

\paragraph{Scene generation and editing.}
Work on generating indoor environments spans text-to-scene pipelines, floorplan synthesis, and procedural layout methods. Recent agentic systems such as SceneSmith~\cite{scenesmith2026} hierarchically compose simulation-ready indoor scenes from natural language through designer--critic--orchestrator agent interactions over stages of architectural layout, furniture, and small-object placement, combining text-to-3D asset synthesis with physics-based feasibility post-processing to produce dense, statically stable rooms with explicit physical properties (mass, friction, inertia, articulation) for direct robotics simulation. EnvDreamer takes a complementary stance focused on corpus-scale embodied training rather than per-scene realism: scenes are interactive and physics enabled, placements obey commonsense and physical constraints, appearance and lighting can vary on load, and tasks compile to executable programs that are verified by validators before inclusion in the released dataset, supporting downstream world-model and embodied policy pretraining.

\paragraph{Robotics simulators and sim to real.}
For robot learning, simulators place greater weight on sensor models, accurate dynamics, and integration with control stacks. Isaac Sim~\cite{isaacsim} provides high fidelity physics, photoreal rendering, and native links to robotics middleware, which makes it well suited for manipulation and perception studies, though scene construction is often manual and not language first. RLBench~\cite{rlbench2020}, RoboSuite~\cite{robosuite2020}, and Sapien~\cite{sapien2020} focus on fine grained manipulation skills and benchmark tasks with detailed kinematics and contact dynamics. EnvDreamer is complementary: it provides language driven scene and task synthesis in Unreal Engine 5 with validator checks and exposes APIs for navigation and manipulation so that policies trained in EnvDreamer can be transferred to real hardware, with image conditioned reconstruction available for sim to real studies when needed.

\section{EnvDreamer}
\label{sec:envdreamer}

\begin{figure*}[t]
    \centering
    \includegraphics[width=\textwidth]{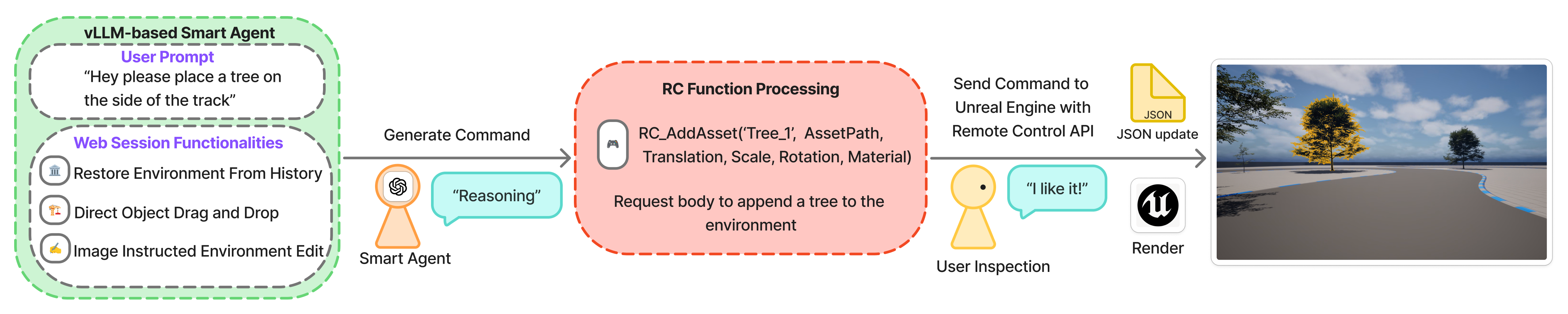}
    \caption{\textbf{Interactive prompt-driven environment editing with EnvDreamer.}
    A GPT-5–based vLLM-based smart agent receives natural-language chat prompts from the user to constantly edit the environment,
    translates them into EnvDreamer API calls, and applies edits
    to a running UE5 scene over the web. Users can inspect the updated environment,
    accept or refine the changes, and leverage additional web-session tools such as
    restoring prior states, direct drag-and-drop editing, and image-instructed edits,
    while the system maintains a synchronized JSON representation of the scene.}
    \label{fig:web_session_overview}
\end{figure*}

EnvDreamer is a framework for language driven generation of interactive Unreal Engine~5 ~\cite{ue5_2024} environments.
It integrates large language and vision language models ~\cite{clip2021} ~\cite{gpt4_2023},
including a GPT-5–class vLLM backend for interactive editing, with the UE5 Remote Control and Python interfaces so that scenes can be authored, edited, and validated at scale.
Figure~\ref{fig:web_session_overview} illustrates an interactive web-session setup in which a vLLM-based smart agent receives user prompts and issues Remote Control commands to modify a running UE5 environment.
As in hand authored UE5 levels, environments produced by EnvDreamer support navigation, manipulation, and multi agent interaction with accurate physics and rich materials.

\paragraph{Overview.}
Given a specification in natural language or a small set of images, EnvDreamer plans assets, layout, and task logic, compiles the plan to Unreal Python, applies changes through Remote Control, and returns renders and a compact scene graph.
A validator suite checks playability, physics consistency, and goal completeness before a scene or task is accepted.
The system exposes a web interface for authoring and review and supports rapid iteration.

\paragraph{Plan–compile–apply–validate loop.}
Figure~\ref{fig:system_overview} illustrates the generation loop.
A planner proposes either a scene program (assets, placements, relations) or a layout description (rooms, bounds, cameras, materials).
A normalizer aligns proposals to a lightweight mirror of the scene graph.
A compiler emits Unreal Python for immediate edits and falls back to C++ only when custom rules, sensors, or physics components are required.
An executor applies actions via Remote Control, captures renders and a scene graph diff, and the validators assess reachability, collisions, articulation constraints, and task predicates.
The loop supports two modes: \emph{direct build}, which applies edits live to the running level, and \emph{layout build}, which materializes reusable layouts into scenes.


\paragraph{Diversity.}
EnvDreamer produces diversity across multiple facets relevant to embodied evaluation.
\emph{Layouts:} from single rooms to multi room houses and outdoor lots, with variable room counts, aspect ratios, and connectivity patterns (Figure~\ref{fig:floorplan_diversity}).
\emph{Assets:} a large library of household, office, and outdoor objects with articulated parts and semantic categories; placement rules reflect commonsense room and location preferences.
\emph{Materials and lighting:} materials, textures, and lighting parameters vary across loads to simulate time of day and appearance changes.
\emph{Behaviors and tasks:} tasks range from object search and rearrangement to tool use and multi step manipulation, encoded as executable programs with explicit preconditions and success metrics ~\cite{metaworld2019}.
\begin{figure}[t]
  \centering
  \includegraphics[width=\linewidth]{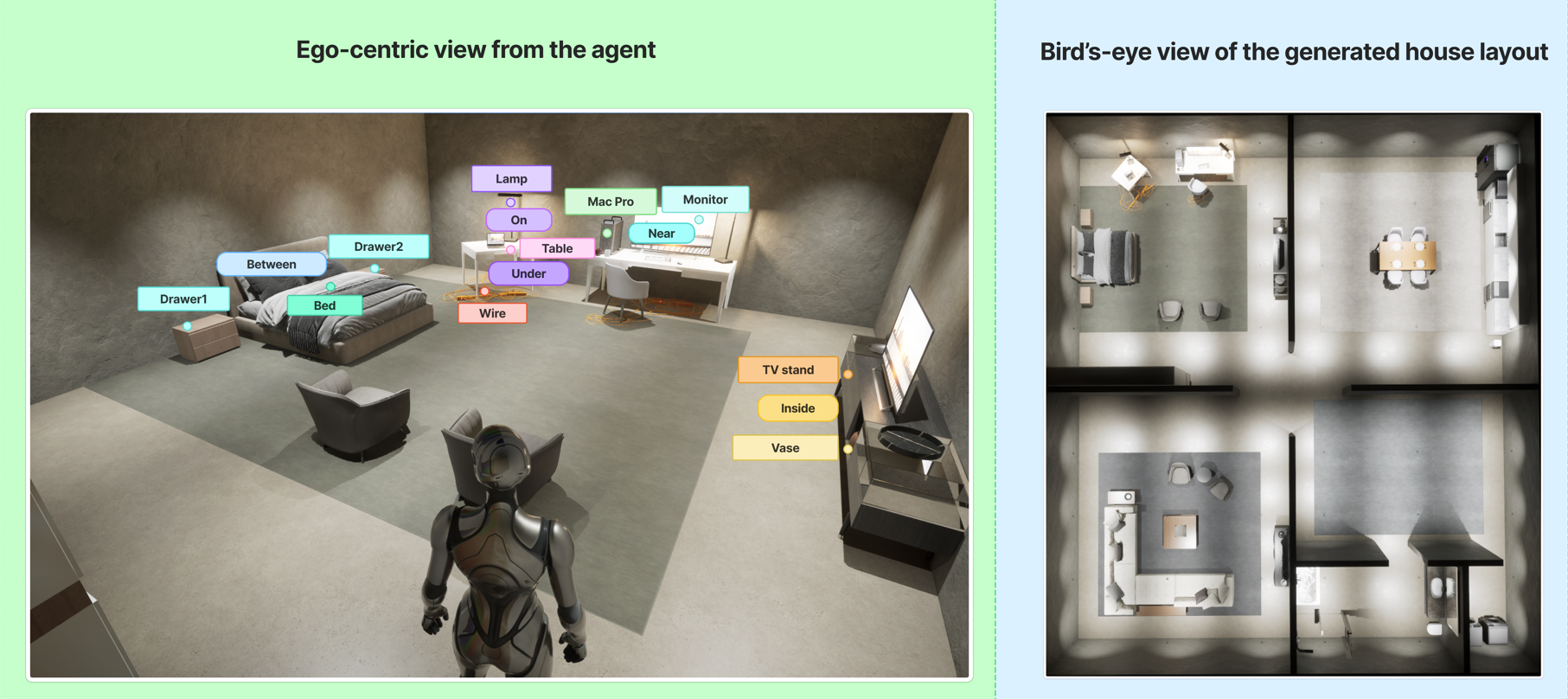}
  \caption{\textbf{Interactive scene in EnvDreamer.}
  Ego-centric view from the embodied agent with annotated object relations
  (\emph{on, under, between, near, inside}) and a bird’s-eye view of the
  procedurally generated multi-room layout.
  The robot is rendered in-scene, and objects are spawned with realistic and diverse geometry, materials, and placements that afford navigation and manipulation.}
  \label{fig:pov_layout}
\end{figure}

\paragraph{Interactivity.}
Objects in EnvDreamer environments support comprehensive physical interactions. Items can be grasped, placed, opened, closed, toggled between states, or articulated through their joints. UE5's physics engine enforces realistic constraints, collisions, and dynamics, enabling complex behaviours such as stacking, tool use, and multi-step manipulation. Appliances have functional states (e.g., ovens that open and heat), containers can be filled or emptied, and furniture can be rearranged. Scenes are authored to preserve navigable free space while maximizing interaction opportunities, with objects placed at reachable heights and orientations that afford natural manipulation. This enables long-horizon tasks requiring sequential interactions with multiple objects capabilities that static scan-based environments cannot provide.

\paragraph{Customization.}
Scene specifications admit simple controls over room counts, categories present or absent, clutter level, lighting ranges, articulation availability, and task families. Small prompt changes or parameter sweeps yield families of related scenes, facilitating controlled generalization tests and curriculum-style progression over layouts and object configurations, as exemplified by the variations in Fig.~\ref{fig:pov_layout}.

\paragraph{Scale and efficiency.}
EnvDreamer represents scenes as compact JSON-like programs with deterministic seeds, enabling both static datasets and on-the-fly generation during training. The plan–compile–apply–validate loop is fully automatic and returns short iterations suitable for large-scale sampling. Scenes load directly in UE5 at training time, keeping storage overhead modest while maintaining high frame rates.

\paragraph{Embodied robots.}
EnvDreamer ships with ready-to-use robot configurations for manipulation and locomotion; Fig.~\ref{fig:pov_layout} shows one humanoid agent in a bedroom configuration. We include a Franka arm for pick-and-place, tool use, and contact-rich tasks, and a Unitree Go1 for exploration, navigation, and retrieval. Policies can access navigation and manipulation interfaces, RGB and depth observations, semantic and instance masks, and domain randomization over textures, lighting, clutter, and physics.

\paragraph{Image to environment for sim to real.}
For sim to real studies, EnvDreamer supports image conditioned generation.
Given photographs of a target space, a vision language model proposes a scene graph aligned to the images, and EnvDreamer instantiates semantically matched assets in UE5.
This capability enables reconstruction of target environments in simulation with minimal manual authoring.

\section{EnvDreamer\textendash 20k}
\label{sec:envdreamer100k}

\begin{figure*}[!t]
    \centering
    \includegraphics[width=\textwidth]{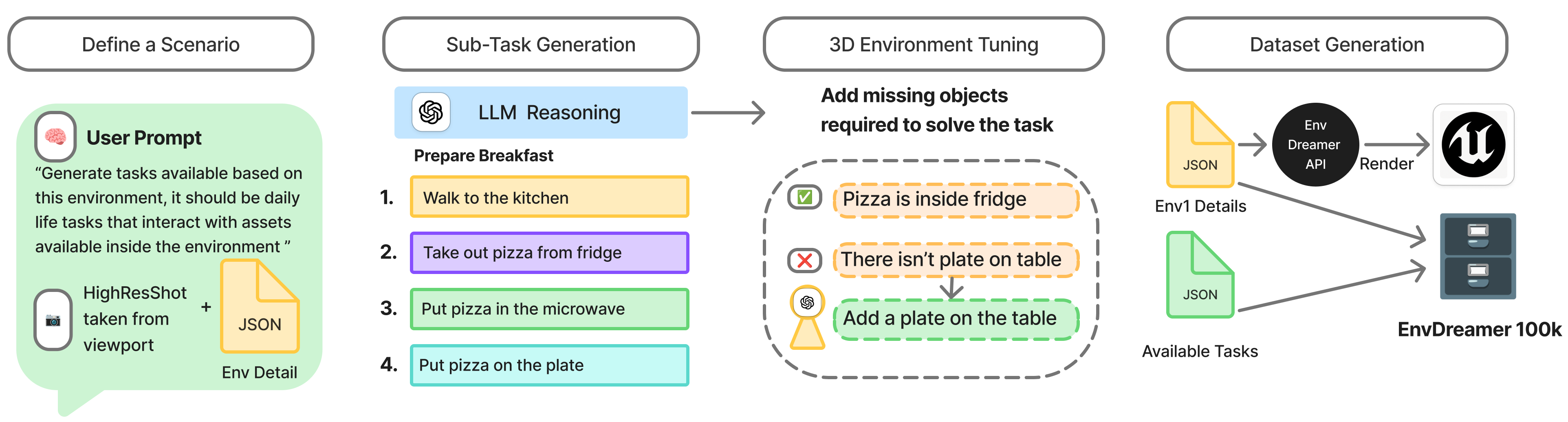}
    \caption{\textbf{Task-conditioned dataset generation pipeline for EnvDreamer\textendash 20k.}
    A user specifies a scenario by providing a prompt and a high-resolution viewport
    screenshot plus the current environment JSON description (left). A language model decomposes
    the request into a sequence of grounded sub-tasks such as ``Walk to the kitchen'' or
    ``Put pizza on the plate'' (middle). The system then checks
    which preconditions are satisfied by comparing object relations and automatically adding missing objects required to solve
    each task (e.g., placing a plate on the table). Finally, the validated environment details and
    the set of available tasks are exported as JSON and rendered through the EnvDreamer API to
    produce entries in the EnvDreamer\textendash 20k dataset (right).}
    \label{fig:ed100k_pipeline}
\end{figure*}

We demonstrate the utility of our platform with a sampled set of 20{,}000 fully interactive environments produced by the generation loop in Section~\ref{sec:envdreamer}, which we refer to as \textbf{EnvDreamer\textendash 20k}. Figure~\ref{fig:ed100k_pipeline} summarizes the task-conditioned pipeline used to convert prompts and environment snapshots into validated scenes and executable task programs. Unless noted otherwise, experiments use a 16k/2k/2k split for train/validation/test; asset and task program splits are detailed in the Appendix. All scenes are fully navigable and support interaction and manipulation without additional authoring. In scale and interactivity, EnvDreamer\textendash 20k is among the largest public corpora designed for embodied AI, world models, and robotics. Scaling beyond 20k is straightforward because each scene is specified by a compact program with deterministic seeds and can be generated on demand during training. Figure~\ref{fig:pov_layout} shows ego centric and top down views.

\begin{figure*}[!t]
    \centering
    \begin{subfigure}[b]{0.32\textwidth}
        \centering
        \includegraphics[width=\linewidth]{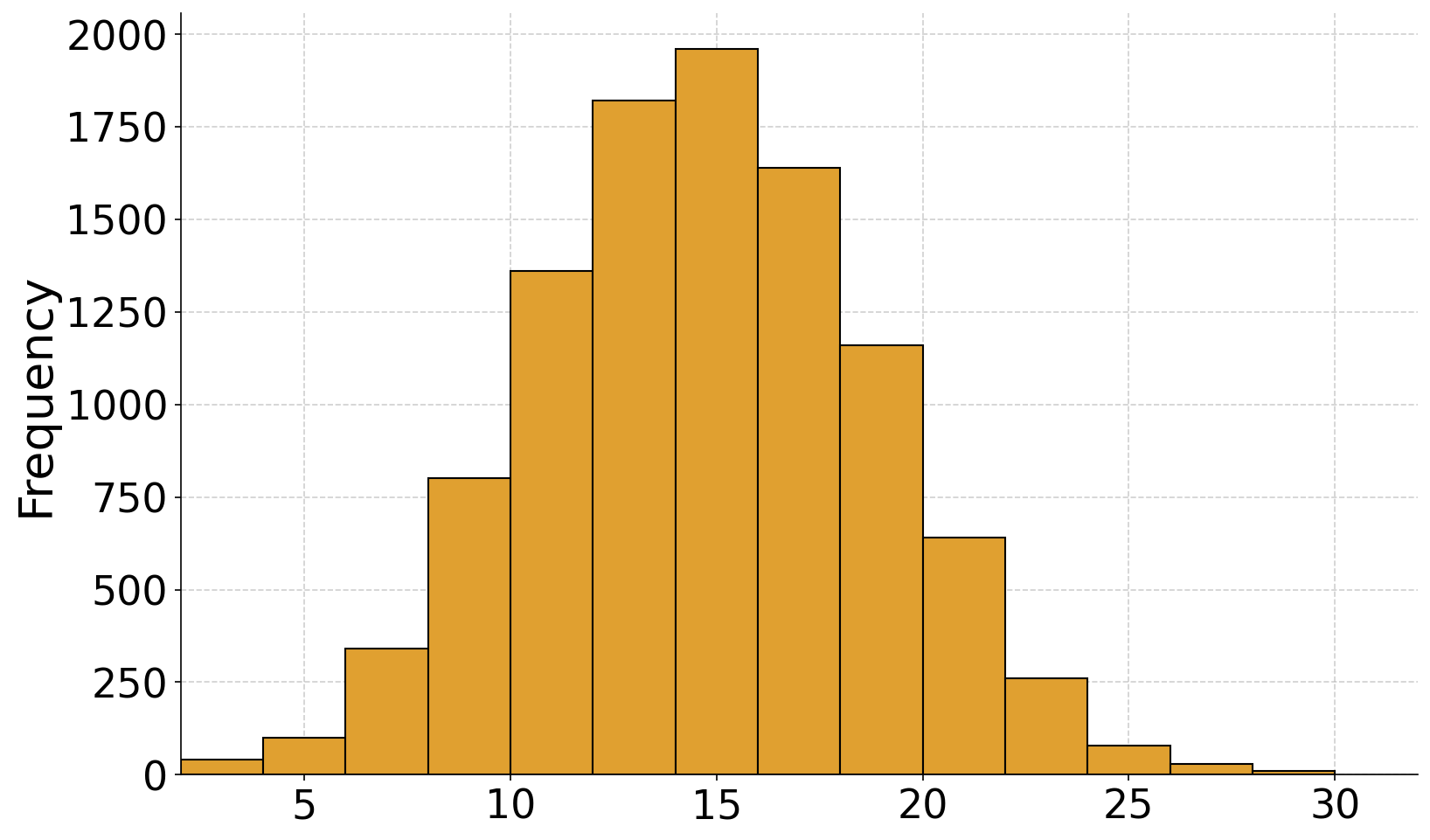}
        \caption{Object counts per room.}
        \label{fig:ed100k_objdist}
    \end{subfigure}
    \hfill
    \begin{subfigure}[b]{0.32\textwidth}
        \centering
        \includegraphics[width=\linewidth]{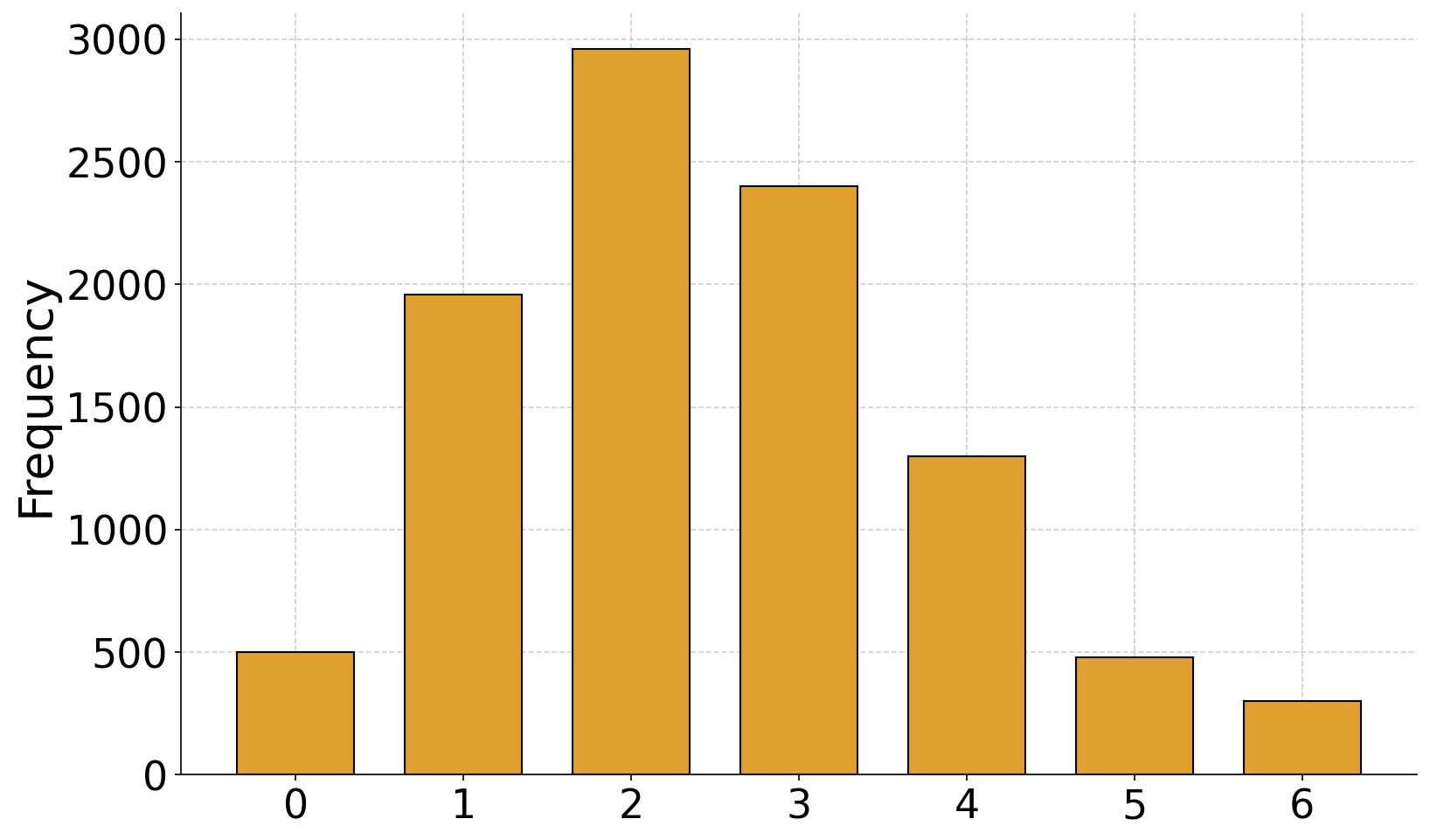}
        \caption{Tasks generated per room.}
        \label{fig:ed100k_taskdist}
    \end{subfigure}
    \hfill
    \begin{subfigure}[b]{0.32\textwidth}
        \centering
        \includegraphics[width=\linewidth]{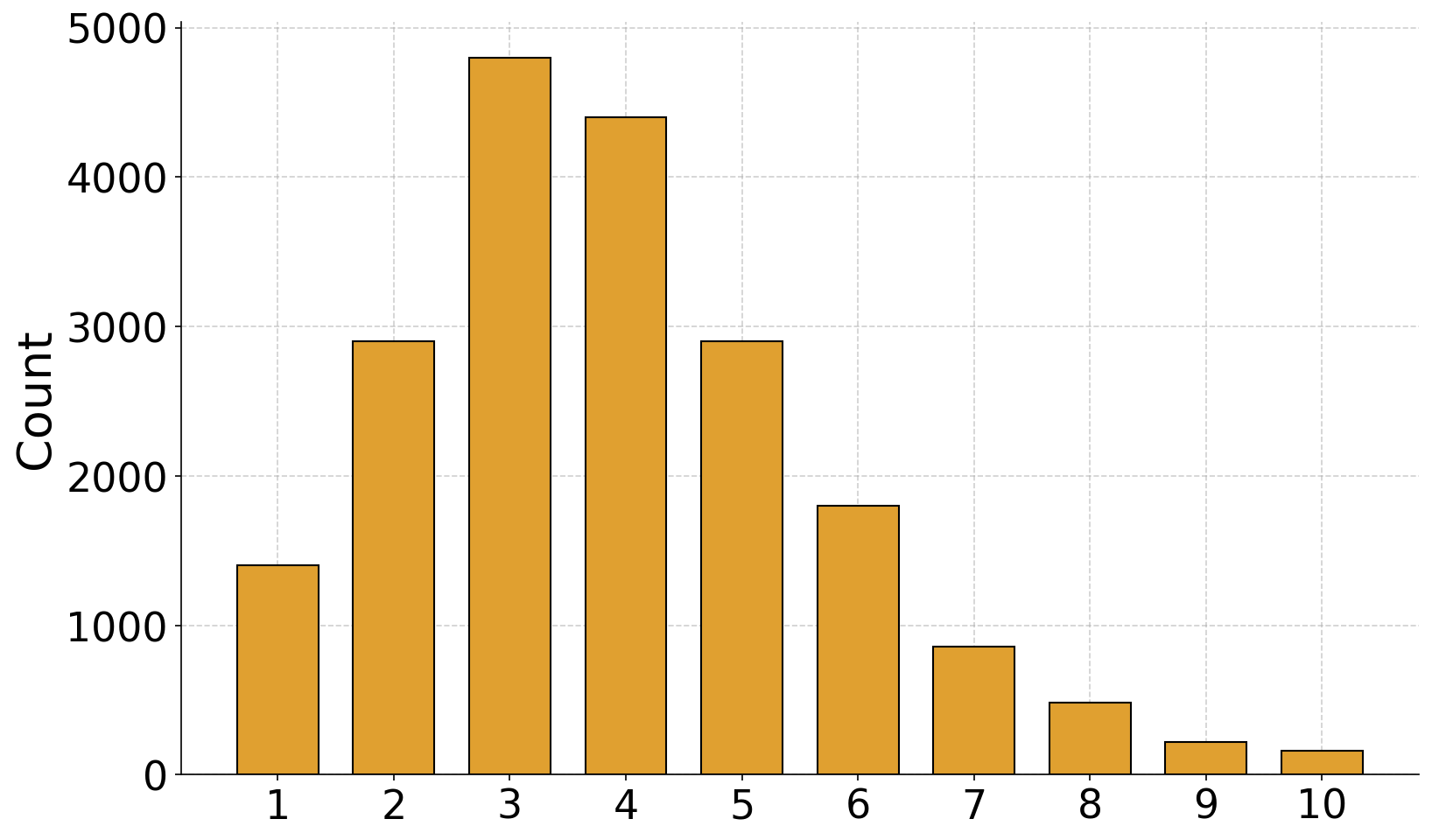}
        \caption{Room counts per environment.}
        \label{fig:ed100k_roomdist}
    \end{subfigure}

    \caption{\textbf{EnvDreamer\textendash20k statistics.}
    \textbf{Left:} distribution of the number of objects per room; most rooms contain 8–18 objects, with a modest tail to 30+ for cluttered spaces. 
    \textbf{Middle:} distribution of the number of tasks generated per room, with the majority of rooms supporting 1–4 grounded tasks and a small fraction supporting 5–6. 
    \textbf{Right:} bar plot of the number of rooms per environment in EnvDreamer\textendash20k, showing that 3–6 room layouts dominate while very small and very large homes are less common.}
    \label{fig:ed100k_stats}
\end{figure*}
\paragraph{Scene statistics.}
EnvDreamer\textendash 20k currently focuses on \emph{indoor} environments such as single rooms, multi-room apartments, and offices.

We control distributions over room count and connectivity via high-level specifications, and empirical statistics are reported in Figure~\ref{fig:ed100k_stats}.
Across the dataset, most environments contain 3--6 rooms, with smaller studio-like scenes and larger multi-room layouts appearing less frequently.
Within each room, object counts follow a controlled distribution centered around 10--16 items, with occasional clutter-heavy rooms containing up to 30 objects.
In addition to structural layout, the system also generates room-conditioned task programs: most rooms support between 1--4 grounded tasks (Figure~\ref{fig:ed100k_stats}, middle), reflecting functional diversity while avoiding overspecified or unrealistic behaviors.
Scenes draw from a broad set of categories spanning furniture, containers, tools, appliances, structural elements, and fixtures.
Placement follows room-type preferences to preserve free space for navigation and afford contact-rich manipulation.
Materials, textures, and lighting vary across loads to simulate different times of day and appearance conditions.


\paragraph{Tasks and validators.}
Each scene includes one or more executable task programs with explicit preconditions and success predicates.
Tasks span object search, waypoint navigation, pick and place and rearrangement, tool use, and multi step manipulation.
Unit style validators check playability, physics consistency, and goal completeness before inclusion.
We report task counts and pass rates per family in Figure~\ref{fig:ed100k_taskdist} and provide failure taxonomies in the Appendix.

\paragraph{Release format.}
Each entry in EnvDreamer\textendash 20k includes: the original prompt or authoring images, the scene program, the scene graph, rendering metadata, task programs and validator outcomes, and optional reference trajectories.
We also provide seed files for exact regeneration and scripts for batch materialization inside Unreal Engine~5.
This format reduces setup cost and supports reproducible benchmarking for world models, embodied agents, and robotics.

\section{Experiments}
\label{sec:experiments}

We empirically evaluate EnvDreamer--20k on navigation and manipulation benchmarks that mirror
the ProcTHOR evaluation protocol. Concretely, we focus on ObjectNav-style navigation in several
simulators and on two manipulation-style tasks: ArmPointNav and one-phase Room
Rearrangement. This allows us to directly compare pretraining on EnvDreamer--20k to pretraining
on ProcTHOR-10K using the same task definitions and metrics.

\subsection{Tasks and Benchmarks}

\paragraph{Navigation.}
We consider ObjectNav (navigation towards a specific object category) in five benchmarks:
ProcTHOR-style houses, ArchitecTHOR, RoboTHOR, HM3D, and AI2-iTHOR. Episodes begin
from a random spawn within the environment and specify a target object category. The agent
receives egocentric RGB observations and must navigate to within a small distance of an instance
of the target category while avoiding collisions. Following standard practice, we report success
rate (SR), success weighted by path length (SPL)~\cite{anderson2018evaluation}.

\paragraph{Manipulation and rearrangement.}
We follow the ProcTHOR setup and evaluate on two manipulation-style tasks:

\begin{itemize}
    \item \textbf{ArmPointNav:} the agent controls a robotic arm to move an object from a source
    location to a destination specified in a 3D coordinate frame. Success is measured by the
    fraction of episodes in which the object ends within a threshold radius of the target .
    \item \textbf{One-phase Room Rearrangement:} the agent must move objects or change their
    state to match a target scene configuration. We report standard rearrangement metrics, such as
    success rate under fixed and strict thresholds~\cite{batra2020rearrangement}.
\end{itemize}

EnvDreamer--20k provides the training environments for these tasks via its validator-checked
task programs and object-centric scene graphs, described in Section~\ref{sec:envdreamer100k}. For
downstream evaluation, we follow the official splits and protocols of each benchmark and restrict
ourselves to RGB inputs only. All EnvDreamer performance numbers reported below were obtained on NVIDIA RTX 5090 GPUs; further compute and parallelism details are provided in the Supplementary Material.








\subsection{Comparison to ProcTHOR}
For fair comparison, all models use identical simple RGB-only architectures without explicit mapping or depth sensing. We evaluate in two modes: zero-shot (pretrained only, no downstream training) and fine-tuning (pretrained then adapted on benchmark data).

ProcTHOR-10K demonstrated that large-scale procedural houses plus a simple RGB-only model can
achieve state-of-the-art results across six embodied benchmarks. EnvDreamer-20k follows the
same philosophy but is built in Unreal Engine 5 and tightly integrates language-driven generation,
task programs, and validator checks. Before comparing performance, we first summarize key
dataset-level differences.

\paragraph{Benchmark-level comparison.}
We next compare three training regimes on each downstream benchmark:

\begin{enumerate}
    \item Training from scratch on the benchmark.
    \item Pretraining on ProcTHOR-10K followed by fine-tuning.
    \item Pretraining on EnvDreamer--20k followed by fine-tuning.
\end{enumerate}

All models share the same architecture and are trained under matched environment-step budgets.
Following ProcTHOR, we report SR and SPL for ObjectNav and success metrics for rearrangement
and ArmPointNav. Table~\ref{tab:results_procthor_envdreamer} shows these results.

\begin{table}[t]
\centering
\scriptsize
\setlength{\tabcolsep}{2pt}
\renewcommand{\arraystretch}{0.88}
\resizebox{\linewidth}{!}{%
\begin{tabular}{lccc}
\toprule
Benchmark & Scratch & P-10K & ED-20k \\
\midrule
ProcTHOR ObjNav      & 18.5$\pm$0.8 / 11.8$\pm$0.6 & 31.4$\pm$1.2 / 19.5$\pm$0.9 & 32.7$\pm$1.1 / 20.3$\pm$0.8 \\
ArchitecTHOR ObjNav  & 14.3$\pm$0.7 / 9.2$\pm$0.5  & 29.8$\pm$1.3 / 18.7$\pm$0.8 & 31.2$\pm$1.2 / 19.4$\pm$0.7 \\
RoboTHOR ObjNav      & 28.3$\pm$1.1 / 15.2$\pm$0.7 & 55.0$\pm$1.8 / 23.7$\pm$1.0 & 56.4$\pm$1.7 / 24.2$\pm$0.9 \\
HM3D ObjNav          & 6.8$\pm$0.4 / 4.2$\pm$0.3   & 9.0$\pm$0.5 / 5.5$\pm$0.4   & 9.8$\pm$0.5 / 5.7$\pm$0.4 \\
AI2-iTHOR ObjNav     & 52.1$\pm$1.9 / 38.4$\pm$1.4 & 75.7$\pm$2.1 / 64.4$\pm$1.8 & 76.3$\pm$2.0 / 64.9$\pm$1.7 \\
Room Rearrangement   & 3.2$\pm$0.3                 & 7.4$\pm$0.6                 & 8.1$\pm$0.5 \\
ArmPointNav          & 24.6$\pm$1.3                & 37.9$\pm$1.6                & 39.2$\pm$1.5 \\
\bottomrule
\end{tabular}%
}
\caption{\textbf{Benchmark performance.}
Scratch: trained only on the benchmark; P-10K: ProcTHOR-10K pretraining; ED-20k: EnvDreamer-20k pretraining. All use the same RGB-only backbone and matched training budgets. EnvDreamer-20k shows consistent improvements over ProcTHOR-10K across all benchmarks.}
\label{tab:results_procthor_envdreamer}
\end{table}

\subsection{Scaling and Ablations}

Finally, we study how performance scales with the size and quality of the EnvDreamer--20k pretraining set, following the scale ablation reported for ProcTHOR. We vary the number of training environments and the strictness of validator checks, and we measure downstream ObjectNav performance under a fixed training budget.

\paragraph{Dataset size.}
We train identical policies on subsets of EnvDreamer--20k containing 5k, 10k, 15k,
and 20k environments, and evaluate zero-shot on representative ObjectNav benchmarks.
Table~\ref{tab:scale_ablation_envdreamer} shows these results.

\begin{table}[t]
\centering
\footnotesize
\setlength{\tabcolsep}{2pt}
\resizebox{\linewidth}{!}{
\begin{tabular}{lcccc}
\toprule
\# ED envs & Arch SR/SPL & Robo SR/SPL & HM3D SR/SPL & AI2 SR/SPL \\
\midrule
5k    & 24.2$\pm$0.9 / 15.3$\pm$0.7 & 42.1$\pm$1.4 / 18.9$\pm$0.8 & 7.9$\pm$0.5 / 4.8$\pm$0.3 & 70.3$\pm$1.8 / 58.2$\pm$1.5 \\
10k   & 27.3$\pm$1.0 / 16.9$\pm$0.8 & 47.2$\pm$1.5 / 20.8$\pm$0.9 & 8.4$\pm$0.5 / 5.2$\pm$0.4 & 72.6$\pm$1.9 / 60.9$\pm$1.6 \\
15k   & 30.1$\pm$1.1 / 18.6$\pm$0.9 & 51.9$\pm$1.6 / 22.6$\pm$1.0 & 9.0$\pm$0.5 / 5.6$\pm$0.4 & 74.8$\pm$1.9 / 63.1$\pm$1.6 \\
20k   & 32.9$\pm$1.2 / 20.3$\pm$0.9 & 56.8$\pm$1.7 / 24.5$\pm$1.1 & 9.7$\pm$0.6 / 5.9$\pm$0.4 & 76.9$\pm$2.0 / 65.2$\pm$1.7 \\
\bottomrule
\end{tabular}
}
\caption{\textbf{Dataset size ablation.}
Policies are pretrained on subsets of EnvDreamer-20k (5k, 10k, 15k, and the full 20k environments) and evaluated zero-shot on four ObjectNav benchmarks. Results show consistent improvements with scale, validating our decision to generate 20k environments.}
\label{tab:scale_ablation_envdreamer}
\end{table}

A paired ablation isolating per-environment quality from raw scale, together with a study of validator strictness, is reported in the Supplementary Material (Section~\ref{sec:supp_ablations}).
\section{Discussion and Limitations}
While EnvDreamer-20k advances the scale of interactive embodied AI environments, several limitations suggest future directions.

\paragraph{Scope and coverage.}
EnvDreamer-20k currently focuses on indoor environments; outdoor scenes are supported by the framework and will be included in future releases. The system lacks complex physics (liquids, deformables) and supports only single-agent tasks. Generation takes 2--5 minutes per environment, and CPU bottlenecks reduce per-instance throughput at high parallelism, though aggregate performance suffices for training.

\paragraph{Task complexity.}
Tasks are limited to relatively short horizons (10--50 steps) with single solutions. Future work will extend to longer-horizon, open-ended specifications with multiple valid approaches that better reflect real-world complexity.

\paragraph{Future directions.}
Key extensions include outdoor environments, advanced physics (liquids, deformables), multi-agent validators, curriculum learning, longer-horizon tasks with multiple solution paths, and a pipeline under development for exporting scenes to NVIDIA Omniverse for use within Isaac Sim and the broader Omniverse robotics tooling. The open-source release enables community extensions to assets, tasks, and validators.

Despite these limitations, EnvDreamer-20k demonstrates that language-driven, validator-checked generation produces diverse, high-quality training data at scale, achieving competitive performance across benchmarks while maintaining rapid iteration capabilities.
\section{Conclusion}

We presented EnvDreamer, a language-first framework for generating arbitrarily large sets of interactive, physics-enabled Unreal Engine~5 environments for embodied AI, world models, and robotics. EnvDreamer-20k—our dataset of 20,000 validated environments with executable task programs, scene graphs, and metadata—enables policies that consistently outperform ProcTHOR-10K pretraining across all tested navigation, rearrangement, and manipulation benchmarks, with strong zero-shot transfer capabilities. Scaling experiments confirm that performance continues to improve with dataset size, validating the investment in large-scale generation. The open-source release of EnvDreamer and the full EnvDreamer-20k dataset aims to democratize access to high-quality training environments and accelerate progress in embodied AI research. By bridging vision-language models with physics simulation and unit-style validation, EnvDreamer provides a scalable, reproducible foundation for embodied intelligence.

\bibliographystyle{plainnat}
\bibliography{references}

@article{anderson2018evaluation,
  title={On Evaluation of Embodied Navigation Agents},
  author={Anderson, Peter and Chang, Angel and Chaplot, Devendra Singh and Dosovitskiy, Alexey and Gupta, Saurabh and Koltun, Vladlen and Kosecka, Jana and Malik, Jitendra and Mottaghi, Roozbeh and Savva, Manolis and Zamir, Amir R.},
  journal={arXiv preprint arXiv:1807.06757},
  year={2018}
}

@article{batra2020rearrangement,
  title={Rearrangement: A Challenge for Embodied {AI}},
  author={Batra, Dhruv and Chang, Angel X. and Chernova, Sonia and Davison, Andrew J. and Deng, Jia and Koltun, Vladlen and Levine, Sergey and Malik, Jitendra and Mordatch, Igor and Mottaghi, Roozbeh and Savva, Manolis and Su, Hao},
  journal={arXiv preprint arXiv:2011.01975},
  year={2020}
}

@inproceedings{clip2021,
  title={Learning Transferable Visual Models from Natural Language Supervision},
  author={Radford, Alec and Kim, Jong Wook and Hallacy, Chris and Ramesh, Aditya and Goh, Gabriel and Agarwal, Sandhini and Sastry, Girish and Askell, Amanda and Mishkin, Pamela and Clark, Jack and Krueger, Gretchen and Sutskever, Ilya},
  booktitle={International Conference on Machine Learning (ICML)},
  year={2021}
}

@article{DBLP:journals/corr/abs-1712-05474,
  title={{AI2-THOR}: An Interactive {3D} Environment for Visual {AI}},
  author={Kolve, Eric and Mottaghi, Roozbeh and Han, Winson and VanderBilt, Eli and Weihs, Luca and Herrasti, Alvaro and Gordon, Daniel and Zhu, Yuke and Gupta, Abhinav and Farhadi, Ali},
  journal={arXiv preprint arXiv:1712.05474},
  year={2017}
}

@inproceedings{DBLP:journals/corr/abs-1806-07011,
  title={{VirtualHome}: Simulating Household Activities via Programs},
  author={Puig, Xavier and Ra, Kevin and Boben, Marko and Li, Jiaman and Wang, Tingwu and Fidler, Sanja and Torralba, Antonio},
  booktitle={IEEE Conference on Computer Vision and Pattern Recognition (CVPR)},
  year={2018}
}

@inproceedings{DBLP:journals/corr/abs-1904-01201,
  title={{Habitat}: A Platform for Embodied {AI} Research},
  author={Savva, Manolis and Kadian, Abhishek and Maksymets, Oleksandr and Zhao, Yili and Wijmans, Erik and Jain, Bhavana and Straub, Julian and Liu, Jia and Koltun, Vladlen and Malik, Jitendra and Parikh, Devi and Batra, Dhruv},
  booktitle={IEEE/CVF International Conference on Computer Vision (ICCV)},
  year={2019}
}

@inproceedings{deitke2022procthor,
  title={{ProcTHOR}: Large-Scale Embodied {AI} Using Procedural Generation},
  author={Deitke, Matt and VanderBilt, Eli and Herrasti, Alvaro and Weihs, Luca and Salvador, Jordi and Ehsani, Kiana and Han, Winson and Kolve, Eric and Farhadi, Ali and Kembhavi, Aniruddha and Mottaghi, Roozbeh},
  booktitle={Advances in Neural Information Processing Systems (NeurIPS)},
  year={2022}
}

@inproceedings{deitke2023phone2proc,
  title={{Phone2Proc}: Bringing Robust Robots into Our Chaotic World},
  author={Deitke, Matt and Hendrix, Rose and Farhadi, Ali and Ehsani, Kiana and Kembhavi, Aniruddha},
  booktitle={IEEE/CVF Conference on Computer Vision and Pattern Recognition (CVPR)},
  year={2023}
}

@techreport{gpt4_2023,
  title={{GPT-4} Technical Report},
  author={{OpenAI}},
  institution={OpenAI},
  year={2023},
  note={arXiv:2303.08774}
}

@inproceedings{housegan2020,
  title={{House-GAN}: Relational Generative Adversarial Networks for Graph-Constrained House Layout Generation},
  author={Nauata, Nelson and Chang, Kai-Hung and Cheng, Chin-Yi and Mori, Greg and Furukawa, Yasutaka},
  booktitle={European Conference on Computer Vision (ECCV)},
  year={2020}
}

@inproceedings{igibson2021,
  title={{iGibson} 1.0: A Simulation Environment for Interactive Tasks in Large Realistic Scenes},
  author={Shen, Bokui and Xia, Fei and Li, Chengshu and Mart{\'\i}n-Mart{\'\i}n, Roberto and Fan, Linxi and Wang, Guanzhi and P{\'e}rez-D'Arpino, Claudia and Buch, Shyamal and Srivastava, Sanjana and Tchapmi, Lyne and Tchapmi, Micael and Vainio, Kent and Fei-Fei, Li and Savarese, Silvio},
  booktitle={IEEE/RSJ International Conference on Intelligent Robots and Systems (IROS)},
  year={2021}
}

@misc{isaacsim,
  title={{NVIDIA} {Isaac Sim}: Robotics Simulation and Synthetic Data Generation},
  author={{NVIDIA}},
  year={2023},
  howpublished={\url{https://developer.nvidia.com/isaac/sim}}
}

@inproceedings{li2021pgdrive,
  title={{MetaDrive}/{PGDrive}: Composing Diverse Driving Scenarios for Generalizable Reinforcement Learning},
  author={Li, Quanyi and Peng, Zhenghao and Feng, Lan and Zhang, Qihang and Xue, Zhenghai and Zhou, Bolei},
  booktitle={IEEE Transactions on Pattern Analysis and Machine Intelligence (TPAMI)},
  year={2022},
  note={arXiv:2109.12674}
}

@inproceedings{li2024behavior1khumancenteredembodiedai,
  title={{BEHAVIOR}-1K: A Human-Centered, Embodied {AI} Benchmark with 1,000 Everyday Activities and Realistic Simulation},
  author={Li, Chengshu and Zhang, Ruohan and Wong, Josiah and Gokmen, Cem and Srivastava, Sanjana and Mart{\'\i}n-Mart{\'\i}n, Roberto and Wang, Chen and Levine, Sergey and Lingelbach, Michael and Sun, Jiankai and others},
  booktitle={Conference on Robot Learning (CoRL)},
  year={2024}
}

@inproceedings{manipulathor2021,
  title={{ManipulaTHOR}: A Framework for Visual Object Manipulation},
  author={Ehsani, Kiana and Han, Winson and Herrasti, Alvaro and VanderBilt, Eli and Weihs, Luca and Kolve, Eric and Kembhavi, Aniruddha and Mottaghi, Roozbeh},
  booktitle={IEEE/CVF Conference on Computer Vision and Pattern Recognition (CVPR)},
  year={2021}
}

@inproceedings{matterport3d2017,
  title={{Matterport3D}: Learning from {RGB-D} Data in Indoor Environments},
  author={Chang, Angel and Dai, Angela and Funkhouser, Thomas and Halber, Maciej and Niessner, Matthias and Savva, Manolis and Song, Shuran and Zeng, Andy and Zhang, Yinda},
  booktitle={International Conference on 3D Vision (3DV)},
  year={2017}
}

@inproceedings{metaworld2019,
  title={{Meta-World}: A Benchmark and Evaluation for Multi-Task and Meta Reinforcement Learning},
  author={Yu, Tianhe and Quillen, Deirdre and He, Zhanpeng and Julian, Ryan and Hausman, Karol and Finn, Chelsea and Levine, Sergey},
  booktitle={Conference on Robot Learning (CoRL)},
  year={2019}
}

@inproceedings{raistrick2024infinigen,
  title={{Infinigen Indoors}: Photorealistic Indoor Scenes using Procedural Generation},
  author={Raistrick, Alexander and Mei, Lingjie and Kayan, Karhan and Yan, Stamatis and Zheng, Yiming and Han, Lahav and Lipson, Hugh and Wang, Yuyan and Stanley, Steven and Lin, David and Chen, Jia Deng},
  booktitle={IEEE/CVF Conference on Computer Vision and Pattern Recognition (CVPR)},
  year={2024}
}

@inproceedings{rlbench2020,
  title={{RLBench}: The Robot Learning Benchmark \& Learning Environment},
  author={James, Stephen and Ma, Zicong and Arrojo, David Rovick and Davison, Andrew J.},
  booktitle={IEEE Robotics and Automation Letters (RA-L)},
  year={2020}
}

@inproceedings{robosuite2020,
  title={{robosuite}: A Modular Simulation Framework and Benchmark for Robot Learning},
  author={Zhu, Yuke and Wong, Josiah and Mandlekar, Ajay and Mart{\'\i}n-Mart{\'\i}n, Roberto and Joshi, Abhishek and Nasiriany, Soroush and Zhu, Yifeng},
  booktitle={arXiv preprint arXiv:2009.12293},
  year={2020}
}

@inproceedings{rplan2019,
  title={Data-Driven Interior Plan Generation for Residential Buildings ({RPLAN})},
  author={Wu, Wenming and Fu, Xiao-Ming and Tang, Rui and Wang, Yuhan and Qi, Yu-Hao and Liu, Ligang},
  booktitle={ACM Transactions on Graphics (SIGGRAPH Asia)},
  year={2019}
}

@inproceedings{sapien2020,
  title={{SAPIEN}: A SimulAted Part-Based Interactive ENvironment},
  author={Xiang, Fanbo and Qin, Yuzhe and Mo, Kaichun and Xia, Yikuan and Zhu, Hao and Liu, Fangchen and Liu, Minghua and Jiang, Hanxiao and Yuan, Yifu and Wang, He and Yi, Li and Chang, Angel X. and Guibas, Leonidas J. and Su, Hao},
  booktitle={IEEE/CVF Conference on Computer Vision and Pattern Recognition (CVPR)},
  year={2020}
}

@inproceedings{scannet2017,
  title={{ScanNet}: Richly-Annotated {3D} Reconstructions of Indoor Scenes},
  author={Dai, Angela and Chang, Angel X. and Savva, Manolis and Halber, Maciej and Funkhouser, Thomas and Niessner, Matthias},
  booktitle={IEEE Conference on Computer Vision and Pattern Recognition (CVPR)},
  year={2017}
}

@inproceedings{sceneformer2021,
  title={{SceneFormer}: Indoor Scene Generation with Transformers},
  author={Wang, Xinpeng and Yeshwanth, Chandan and Niessner, Matthias},
  booktitle={International Conference on 3D Vision (3DV)},
  year={2021}
}

@inproceedings{scenesmith2026,
  title={{SceneSmith}: Agentic Composition of Simulation-Ready Indoor Scenes from Natural Language},
  author={Anonymous},
  booktitle={Under review},
  year={2026},
  note={Placeholder citation}
}

@inproceedings{text2scene2019,
  title={{Text2Scene}: Generating Compositional Scenes from Textual Descriptions},
  author={Tan, Fuwen and Feng, Song and Ordonez, Vicente},
  booktitle={IEEE Conference on Computer Vision and Pattern Recognition (CVPR)},
  year={2019}
}

@misc{ue5_2024,
  title={Unreal Engine 5},
  author={{Epic Games}},
  year={2024},
  howpublished={\url{https://www.unrealengine.com/en-US/unreal-engine-5}}
}

@misc{virtualenv2026,
  title={{VirtualEnv}: A Platform for Embodied {AI} Research},
  author={Kabir Swain and Sijie Han and Ayush Raina and Jin Zhang and Shuang Li and Michael Stopa and Antonio Torralba},
  year={2026},
  eprint={2601.07553},
  archivePrefix={arXiv},
  primaryClass={cs.AI},
  url={https://arxiv.org/abs/2601.07553}
}

@inproceedings{wei2022chainofthought,
  title={Chain-of-Thought Prompting Elicits Reasoning in Large Language Models},
  author={Wei, Jason and Wang, Xuezhi and Schuurmans, Dale and Bosma, Maarten and Ichter, Brian and Xia, Fei and Chi, Ed H. and Le, Quoc V. and Zhou, Denny},
  booktitle={Advances in Neural Information Processing Systems (NeurIPS)},
  year={2022}
}

@inproceedings{yang2024holodeck,
  title={{Holodeck}: Language Guided Generation of {3D} Embodied {AI} Environments},
  author={Yang, Yue and Sun, Fan-Yun and Weihs, Luca and VanderBilt, Eli and Herrasti, Alvaro and Han, Winson and Wu, Jiajun and Haber, Nick and Krishna, Ranjay and Liu, Lingjie and Callison-Burch, Chris and Yatskar, Mark and Kembhavi, Aniruddha and Clark, Christopher},
  booktitle={IEEE/CVF Conference on Computer Vision and Pattern Recognition (CVPR)},
  year={2024}
}

@inproceedings{zhang2024architect,
  title={{Architect}: Generating Vivid and Interactive {3D} Scenes with Hierarchical {2D} Inpainting},
  author={Zhang, Yian and Wang, Xiaoshi and Lyu, Yuyang and Zhao, Wenliang and Liu, Jingdong and Wang, Yinpeng and Yu, Jingyi},
  booktitle={Advances in Neural Information Processing Systems (NeurIPS)},
  year={2024}
}

\clearpage
\section{Supplementary Materials}
\label{sec:supp}

\subsection{EnvDreamer Remote Control API}
\label{sec:supp_api}

EnvDreamer exposes a lightweight Remote Control (RC) API that allows
Python tools and external agents to programmatically construct, inspect,
and manipulate scenes inside Unreal Engine. The API is implemented by the
\texttt{AJsonSceneBuilder} actor on the engine side and a small set of
client utilities on the Python side (e.g., \texttt{rc\_call()} in
\texttt{HomeAgent}). This section summarizes the design and typical
usage patterns of the API.


\begin{figure}[t]
  \centering
  \includegraphics[width=0.9\linewidth]{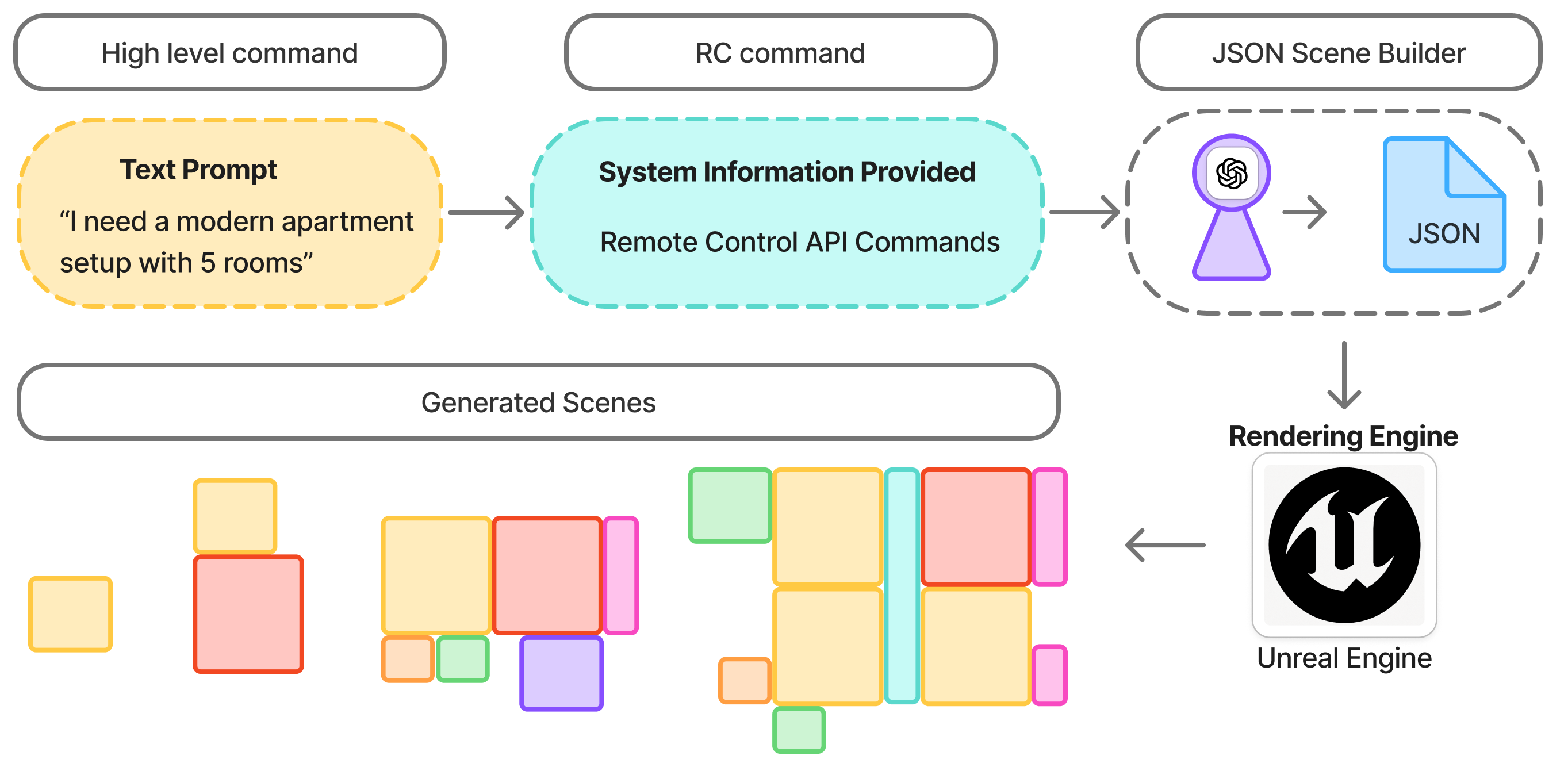}
  \caption{High-level architecture of the EnvDreamer Remote Control API.
  Python clients issue JSON commands to \texttt{AJsonSceneBuilder}, which
  resolves assets, spawns actors, and returns structured responses.}
  \label{fig:rc_architecture}
\end{figure}

\subsubsection{JsonSceneBuilder Actor}

The \texttt{AJsonSceneBuilder} actor is responsible for:
\begin{itemize}
  \item maintaining a map from user-level identifiers (\texttt{Id}) to
        spawned actors using tags of the form \texttt{"RCID:\textless Id\textgreater"},
  \item interpreting RC commands (add/remove assets, set transforms,
        apply materials/textures), and
  \item exposing camera and snapshot utilities for debugging and data
        collection.
\end{itemize}

Existing actors marked as \texttt{"JsonGenerated"} are re-indexed at
startup via \texttt{ReindexExistingJsonActors()}, allowing scenes built in
previous sessions to be reused without re-spawning from scratch.

\subsubsection{Asset Resolution and Registry}

To decouple layout descriptions from engine paths, EnvDreamer uses a
simple asset registry:

\begin{itemize}
  \item \textbf{Config file.}
        At editor startup, \texttt{AssetRegistry::Initialize()} loads
        \texttt{Config/asset\_registry.json}, which stores assets grouped
        by semantic categories (e.g., \texttt{"kitchen"}, \texttt{"office"})
        and an optional alias map.
  \item \textbf{Name resolution.}
        \texttt{AssetRegistry::Resolve(NameOrPath)} returns a fully
        qualified Unreal path when given a shortcut name (e.g.,
        \texttt{"bp\_table"}) or passes through raw \texttt{/Game/...}
        paths unchanged.
  \item \textbf{Type detection.}
        \texttt{RC\_AddAsset} uses Unreal’s \texttt{IAssetRegistry} to
        distinguish between static mesh assets and Blueprint classes,
        falling back to \texttt{"\_C"} suffix heuristics when metadata is
        unavailable.
\end{itemize}

\subsubsection{Core Asset and Material Operations}

Table~\ref{tab:rc_methods} summarizes the most frequently used RC
functions exposed by \texttt{AJsonSceneBuilder}. All commands are
identified by a string \texttt{Id} that serves as a stable handle for
future updates.


\begin{table}[t]
\centering
\caption{Key EnvDreamer RC methods callable via Python.}
\label{tab:rc_methods}
\small
\setlength{\tabcolsep}{4pt}
\begin{tabular}{@{}ll@{}}
\toprule
Method & Purpose \\
\midrule
\texttt{RC\_AddAsset} & Spawn mesh/Blueprint at pose \\
\texttt{RC\_Remove} & Destroy actor by ID \\
\texttt{RC\_SetLocation} & Update translation \\
\texttt{RC\_SetScale} & Update actor scale \\
\texttt{RC\_SetMaterialColor} & Apply colored material \\
\texttt{RC\_SetMaterialTexture} & Apply texture asset \\
\texttt{ClearAllJsonActors} & Remove all generated \\
\bottomrule
\end{tabular}
\vspace{-3mm}
\end{table}

\paragraph{RC\_AddAsset.}
Given an \texttt{Id}, \texttt{AssetPath}, pose, and scale, the engine
first resolves the asset name through the registry, then auto-detects
whether the asset is a static mesh or a Blueprint:

\begin{itemize}
  \item For static meshes, a new \texttt{AStaticMeshActor} is spawned and
        its \texttt{UStaticMeshComponent} is configured (mobility, mesh,
        optional color material).
  \item For Blueprints, \texttt{LoadBlueprintClass} converts the asset
        object path to a class path (appending \texttt{"\_C"} as needed)
        and spawns an instance of that class.
\end{itemize}

In both cases, the resulting actor is tagged with \texttt{"JsonGenerated"}
and \texttt{"RCID:\textless Id\textgreater"}, added to the internal index,
and returned to the client as a success indicator.

\paragraph{Material and texture control.}
Two complementary operations are provided:

\begin{itemize}
  \item \texttt{RC\_SetMaterialColor(Id, Color)} uses a cached
        \texttt{UMaterial} (e.g., \texttt{M\_Basic\_Wall}) to create
        per-color dynamic material instances, keyed by the RGBA tuple.
  \item \texttt{RC\_SetMaterialTexture(Id, TexturePath)} first attempts to
        load a full material at the given path (normalizing redundant
        \texttt{.MaterialName} suffixes). If successful, the material is
        applied directly. Otherwise, the code falls back to loading the
        asset as a \texttt{UTexture2D} and wiring it into a dynamic
        material instance via an existing or common texture parameter
        name (e.g., \texttt{"BaseColor"}, \texttt{"Albedo"}).
\end{itemize}

This dual-path strategy allows the same API to handle both authored
materials and raw textures referenced by the Python side.

\subsubsection{Camera Control and Snapshots}

For inspection and dataset generation, the RC API provides basic camera
and screenshot control (editor-only):

\begin{itemize}
  \item \texttt{RC\_SetCameraLocation}, \texttt{RC\_SetCameraRotation},
        and \texttt{RC\_MoveCameraRelative} directly manipulate the active
        editor viewport via \texttt{FEditorViewportClient}.
  \item \texttt{RC\_Snapshot(Name, ResX, ResY, OutPath)} configures the
        high-resolution screenshot system, clears previous PNGs under
        \texttt{Saved/AIShots/}, and requests a new screenshot with the
        given name and resolution. The absolute path is returned to the
        caller for downstream processing.
\end{itemize}

\subsubsection{Introspection Utilities}

To support higher-level reasoning and placement logic, the API exposes
two introspection methods:

\begin{itemize}
  \item \textbf{\texttt{RC\_GetActorInfo(Id)}:}
        Returns a compact JSON string containing the actor’s class name,
        location, rotation, and scale. This is used by Python tools to
        confirm that layout edits have taken effect and to compute
        relative transforms between objects.
  \item \textbf{\texttt{RC\_GetComponentInfo(Id, OutJson)}:}
        Serializes all components attached to the actor, including per
        component transform, bounding box, and inferred semantic tags
        (e.g., \texttt{"surface"} or \texttt{"floor"} based on name
        heuristics). For surface-like components, the method also reports
        a suggested \texttt{placement\_height} above the top of the
        bounding box, which is used by placement policies to drop new
        objects on tables or floors without hard-coding offsets.
\end{itemize}

Together, these endpoints turn EnvDreamer into an inspection-friendly
environment: external agents can not only spawn and restyle assets, but
also query geometric affordances and build richer internal scene
representations without direct access to engine internals.

\subsection{Remote Unreal Web Session}
\label{sec:supp_web}

To make EnvDreamer accessible without opening the Unreal Editor, a
minimal ``simple builder'' web session is provided. It couples a Pixel
Streaming viewport running inside UE5 with a small FastAPI backend and a
single-page HTML/JavaScript client. This interface is used for quick
qualitative inspection, debugging, and human-in-the-loop layout editing.

\begin{figure}[t]
  \centering
  \includegraphics[width=0.9\linewidth]{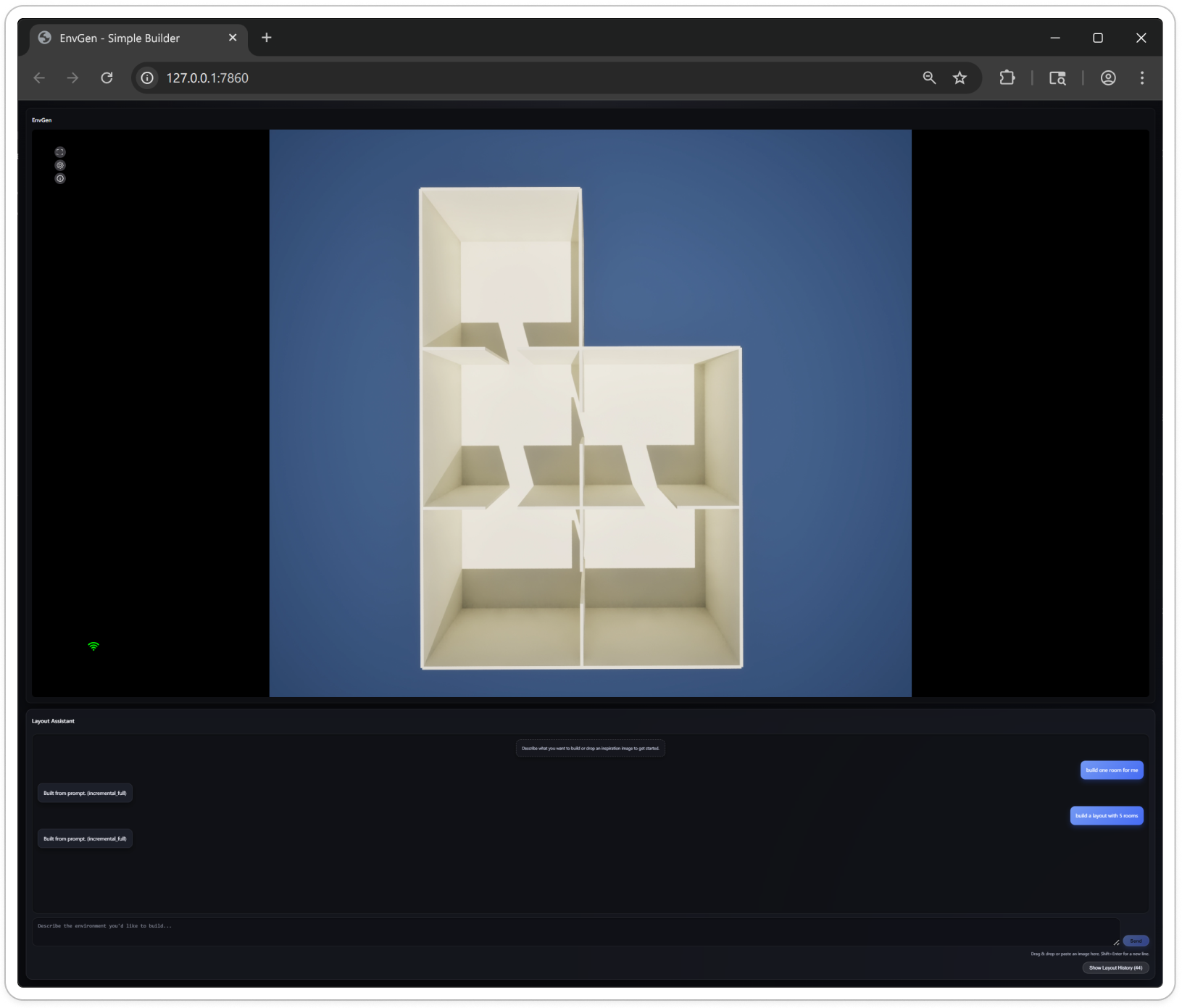}
  \caption{Remote web session for EnvDreamer. The live UE5 viewport is
  streamed to the browser, while a simple builder UI issues layout
  updates through HTTP endpoints backed by the RC API.}
  \label{fig:websession_ui}
\end{figure}

\subsubsection{Pixel-Streamed Viewer}

On the engine side, Unreal’s Pixel Streaming stack exposes the current
editor or game viewport as a low-latency WebRTC stream. The web client
embeds this stream in an \texttt{<iframe>} (pointing to the local Pixel
Streaming address) and forwards mouse and keyboard events directly to
UE5. This allows a remote user to orbit, pan, and fly through the
generated environment and visually confirm that RC-issued layout changes
have taken effect. No simulation logic runs in the browser; the page is a
thin remote viewer and input forwarder.

\subsubsection{Simple Builder Server}

The simple builder backend is a FastAPI application that exposes a small
set of HTTP endpoints:

\begin{itemize}
  \item \textbf{\texttt{GET /}} serves the static HTML page
        (\texttt{web\_simple.html}) containing the viewer and chat-style
        layout assistant.
  \item \textbf{\texttt{GET /shots/latest.png}} returns the latest
        high-resolution screenshot produced by UE5. Screenshots are
        copied into a public directory and served with no-cache headers.
  \item \textbf{\texttt{POST /api/simple/build-from-prompt}} accepts a
        JSON payload with a free-form text description, calls
        \texttt{generate\_layout\_from\_prompt()}, writes the resulting
        layout JSON, and invokes \texttt{apply\_layout\_update()} to
        rebuild the scene.
  \item \textbf{\texttt{POST /api/simple/build-from-image}} accepts an
        uploaded reference image plus an optional text prompt, calls
        \texttt{generate\_layout\_from\_image()}, and applies the
        resulting layout in the same way as the text-only endpoint.
\end{itemize}

Both build endpoints return a compact update summary describing the
change mode (full vs.\ incremental), which rooms were touched, and which
items were updated. The web client uses this summary to render concise
status messages in the chat log.

\subsubsection{Browser UI and Interaction Loop}

The HTML client exposes two main panels: a viewer panel with the embedded
Pixel Streaming iframe, and a ``Layout Assistant'' panel that implements a
chat-style interface. Each user request is echoed as a user message, and
each successful build is summarized as an assistant message. Drag-and-drop
handlers allow a user to paste or drop an image directly into the chat
area, which triggers the image-conditioned endpoint.

The JavaScript logic maintains a small amount of local state, including a
reference count of active HTTP requests (used to disable the send button
and prevent overlapping builds) and error handling that surfaces
non-2xx responses as concise error messages in the chat log. Layout state
is not cached in the browser; the authoritative layout and history live
on disk and are queried when needed.

\subsubsection{Layout History and Checkpoints}

To support backtracking and side-by-side comparisons, the simple builder
records layout checkpoints under \texttt{Config/LayoutHistory/}. Each
successful build creates a JSON snapshot and appends a version entry to
\texttt{history.json} with a timestamp, source type (prompt, image,
clear, etc.), and an optional note. The browser queries
\texttt{GET /api/simple/layout/state}, which returns the current layout,
the list of versions, and the identifier of the active checkpoint; this
list populates the ``Layout History'' sidebar with simple ``Previous'',
``Next'', and ``Restore'' controls.

Dedicated endpoints
(\texttt{/api/simple/layout/restore},
 \texttt{/api/simple/layout/clear},
 \texttt{/api/simple/layout/set-wall-color})
allow a user to restore a previous snapshot, clear the layout, or apply
targeted edits (such as changing a room’s wall color) without re-running
the planner. Restoring a checkpoint updates the default layout file,
invokes \texttt{apply\_layout\_update()} to resynchronize UE5, and marks
the selected version as current in \texttt{history.json}.

\subsubsection{Intended Usage}

The remote web session is intended for rapid manual exploration and
debugging rather than large-scale training. Typical uses include:
interactive testing of planner behaviour without opening the Unreal
Editor, quick qualitative evaluation of layout diversity and failure
modes by non-expert users, and easy reproduction of specific layouts from
the paper by loading the corresponding checkpoint and streaming the
resulting scene. Because the interface is built entirely on top of the RC
API and planner endpoints, it can be deployed on remote hosts or extended
with task-specific controls without changes to the core engine.

\subsection{Layout Planner and Prompt Design}
\label{sec:supp_planner}

The main paper describes the overall EnvDreamer pipeline at a high level.
Here we focus on the layout planner component that translates text (and
optionally images) into editable layouts.

\subsubsection{Planner API and Control Flow}

The planner exposes a small API used by the UE5 tools:

\begin{itemize}
  \item \texttt{generate\_layout\_from\_prompt(prompt, current\_layout)}  
        produces a full layout JSON given a natural-language instruction and,
        optionally, a pre-existing layout to edit.
  \item \texttt{generate\_layout\_from\_image(prompt, image\_bytes, current\_layout)}  
        reconstructs a layout from a reference image plus optional text.
  \item \texttt{validate\_layout\_schema(layout)} and
        \texttt{write\_layout\_file(path, layout)}  
        provide lightweight validation and serialization.
\end{itemize}

Internally, both generation functions construct a list of messages that
combine:
(1) a global system prompt describing geometric and semantic constraints;
(2) an optional summary of the current layout (when editing); and
(3) a strict JSON schema specification that pins down the output format.
The user instruction is appended as the final \texttt{user} message.
We request a structured response via
\texttt{response\_format=\{\,"type": "json\_object"\,\}}, which avoids the
need for ad-hoc text post-processing.

\begin{figure}[t]
  \centering
  \includegraphics[width=0.9\linewidth]{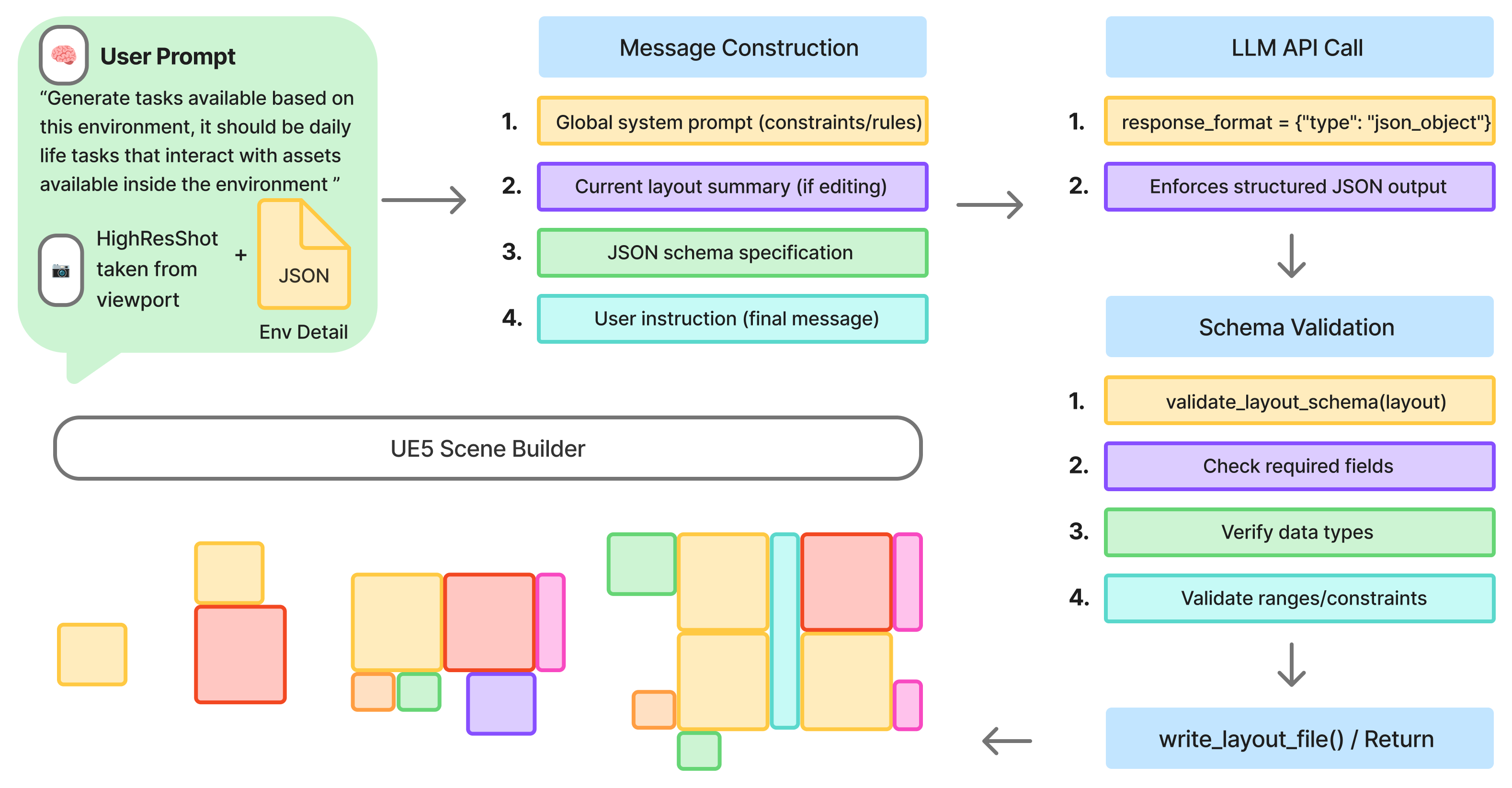}
  \caption{High-level control flow of the layout planner.}
  \label{fig:planner_flow}
\end{figure}

\subsubsection{Incremental Editing Behaviour}

In interactive sessions the user rarely specifies an entire apartment in
one prompt. Instead, EnvDreamer receives a sequence of localized edits:
``extend the kitchen by one grid cell,'' ``replace the sofa with two arm
chairs,'' or ``turn the study into a bedroom.'' To support these
workflows:

\begin{itemize}
  \item The planner always passes the most recent layout as an additional
        \texttt{system} message tagged as \texttt{current\_layout}.
        The prompt emphasizes that unspecified rooms and items should be
        preserved.
  \item When the user request is ambiguous (e.g., ``make the living room
        brighter''), we treat the instruction as a soft preference and
        allow the LLM to choose between adjusting lighting, swapping
        materials, or both. The resulting layout is still filtered by the
        validator.
  \item If schema validation fails, the tool surfaces a concise error
        message and offers to either (i) re-issue the prompt with a
        stronger constraint hint, or (ii) revert to the previous layout.
\end{itemize}

\begin{figure}[t]
  \centering
  \includegraphics[width=0.9\linewidth]{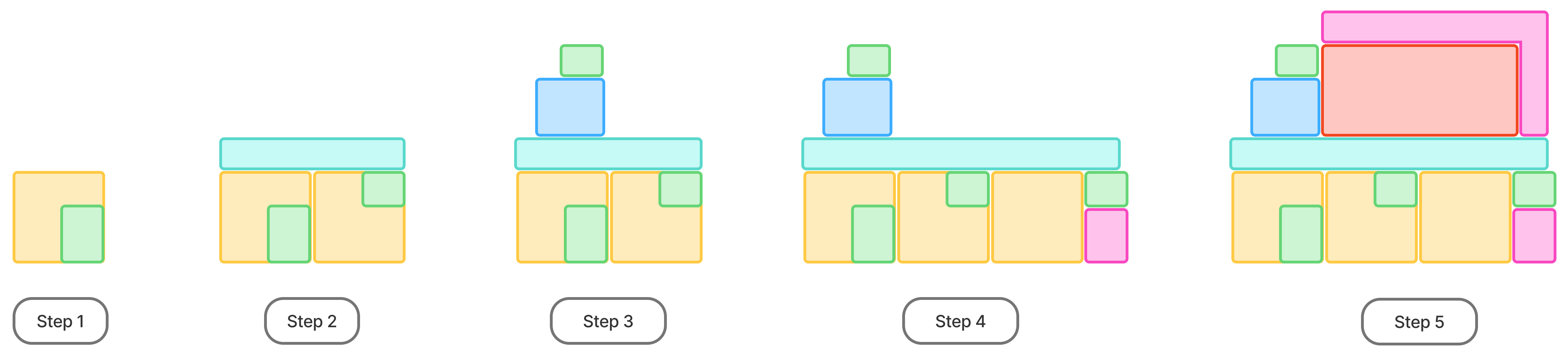}
  \caption{Illustrative incremental editing session.}
  \label{fig:edit_cases}
\end{figure}

\subsubsection{Vision-conditioned Layouts}

For image-conditioned generation, we embed the input photograph as an
\texttt{image\_url} in the \texttt{user} message and prepend a short
instruction such as ``\emph{Infer a plausible floor plan for this room}''
or ``\emph{Align the furniture arrangement with this photo while keeping
the existing topology}.'' Unlike the text-only case, the planner treats
the image as soft evidence: it aims to match visible structure and object
types while still respecting grid snapping and validator constraints.

We found that allowing the model to slightly deviate from the exact pixel
geometry leads to layouts that are easier to navigate and reuse across
tasks, at the cost of a small loss in photometric fidelity.


\subsection{Additional EnvDreamer Dataset Diagnostics}
\label{sec:supp_dataset}

The main paper reports aggregate dataset statistics and task coverage.
Here we provide additional diagnostics that are useful for understanding
what kinds of environments EnvDreamer tends to produce.

\subsubsection{Room Count and Floor Area}

\begin{figure}[t]
  \centering
  \includegraphics[width=0.9\linewidth]{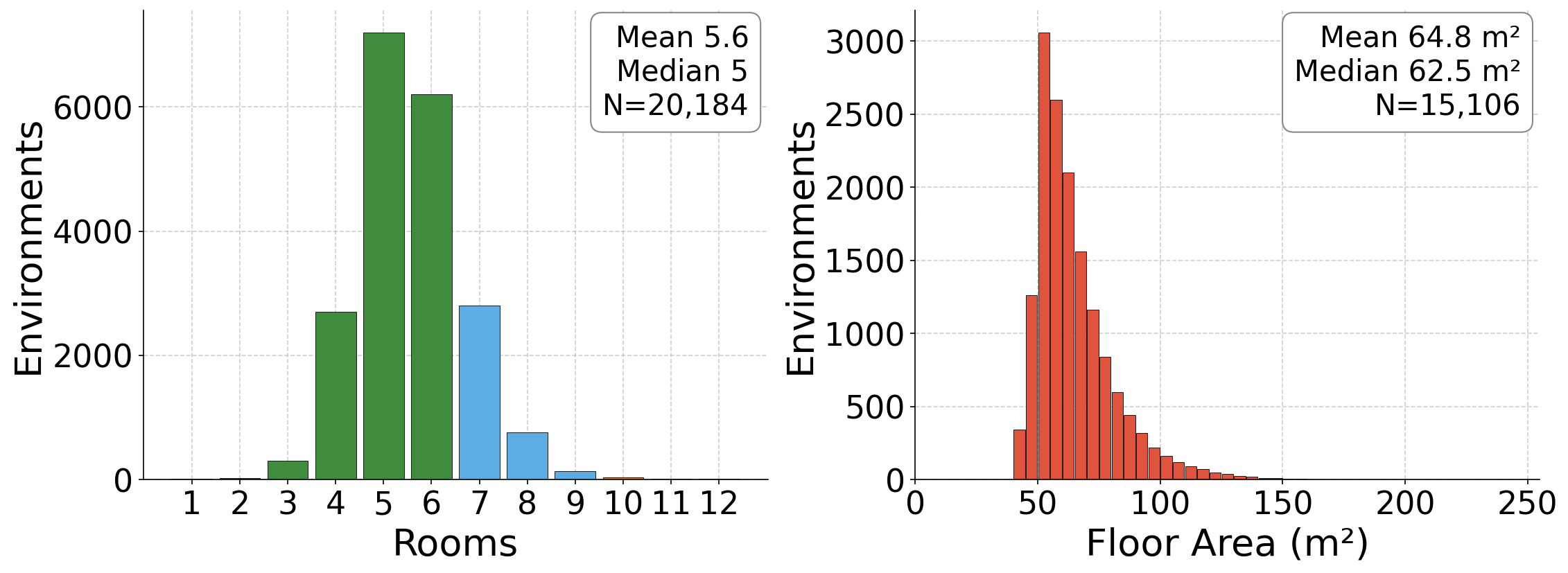}
  \caption{Distribution of room counts and usable floor area.}
  \label{fig:room_hist}
\end{figure}

Figure~\ref{fig:room_hist} sketches the distribution over room counts per
environment and total usable floor area (walkable area minus large
fixtures). While the generator allows single-room studios and large
multi-room apartments, the bulk of the distribution lies between 3 and
6 rooms and 40--120~m$^2$.

\subsubsection{Object Co-occurrence Structure}

\begin{figure}[t]
  \centering
  \includegraphics[width=0.9\linewidth]{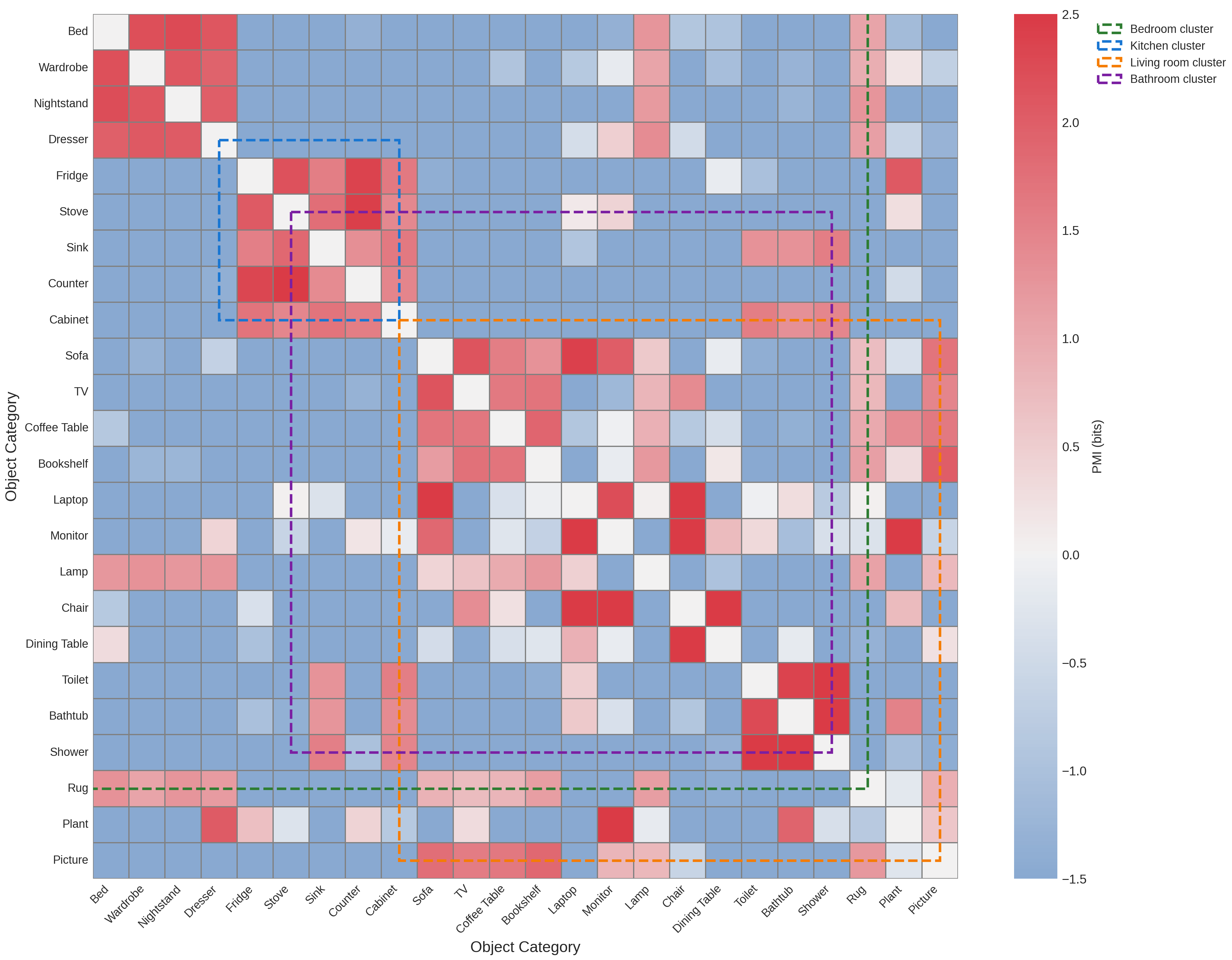}
  \caption{Category-level PMI highlighting typical furniture clusters.}
  \label{fig:cooccur}
\end{figure}

To better understand how objects cluster across rooms, we estimate a
co-occurrence matrix over asset categories and compute pointwise mutual
information (PMI). Figure~\ref{fig:cooccur} reveals:

\begin{itemize}
  \item strong positive associations between beds, wardrobes, and night
        stands in bedrooms;
  \item typical kitchen items involving fridges, stoves, sinks, and
        counters; and
  \item weaker but still positive correlations between electronic devices
        (e.g., laptops, TVs) and seating furniture.
\end{itemize}

These statistics can be used to drive more structured curriculum design
(e.g., progressively increasing the ``clutter'' factor) in future work.

\subsubsection{Layout Diversity over Prompts}

Finally, we examine prompt-level diversity by sampling multiple layouts
for the same high-level description (e.g., ``a compact two-bedroom
apartment for students'') and measuring variation in:
(1) number of rooms,
(2) connectivity structure, and
(3) object counts.
Table~\ref{tab:prompt_diversity} is reserved for these statistics.


\begin{table}[t]
\centering
\caption{Prompt-level diversity: statistics for 16 samples per prompt.}
\label{tab:prompt_diversity}
\small
\begin{tabular}{@{}lccc@{}}
\toprule
Prompt & Rooms & Degree & Objects \\
\midrule
2BR student apt. & $4.4\pm1.3$ & $2.8\pm0.9$ & $29\pm8$ \\
Open studio & $2.2\pm0.5$ & $1.6\pm0.4$ & $21\pm4$ \\
3BR family home & $7.1\pm1.9$ & $3.0\pm0.7$ & $46\pm12$ \\
1BR minimalist & $3.7\pm0.8$ & $2.3\pm0.6$ & $25\pm7$ \\
Luxury penthouse & $7.9\pm2.2$ & $3.2\pm1.0$ & $64\pm15$ \\
Rustic cottage & $5.3\pm1.1$ & $2.6\pm0.5$ & $38\pm10$ \\
Urban loft & $4.1\pm0.7$ & $2.2\pm0.4$ & $32\pm6$ \\
4BR traditional & $8.6\pm1.8$ & $3.1\pm0.8$ & $69\pm11$ \\
Tiny home & $2.6\pm0.6$ & $1.8\pm0.3$ & $19\pm5$ \\
Senior bungalow & $4.5\pm0.9$ & $2.5\pm0.7$ & $34\pm8$ \\
\bottomrule
\end{tabular}
\vspace{-3mm}
\end{table}
\subsection{Robot Configurations}
\label{sec:supp_robots}

\begin{figure}[t]
  \centering
  \begin{subfigure}[b]{0.48\linewidth}
    \centering
    \includegraphics[width=\linewidth]{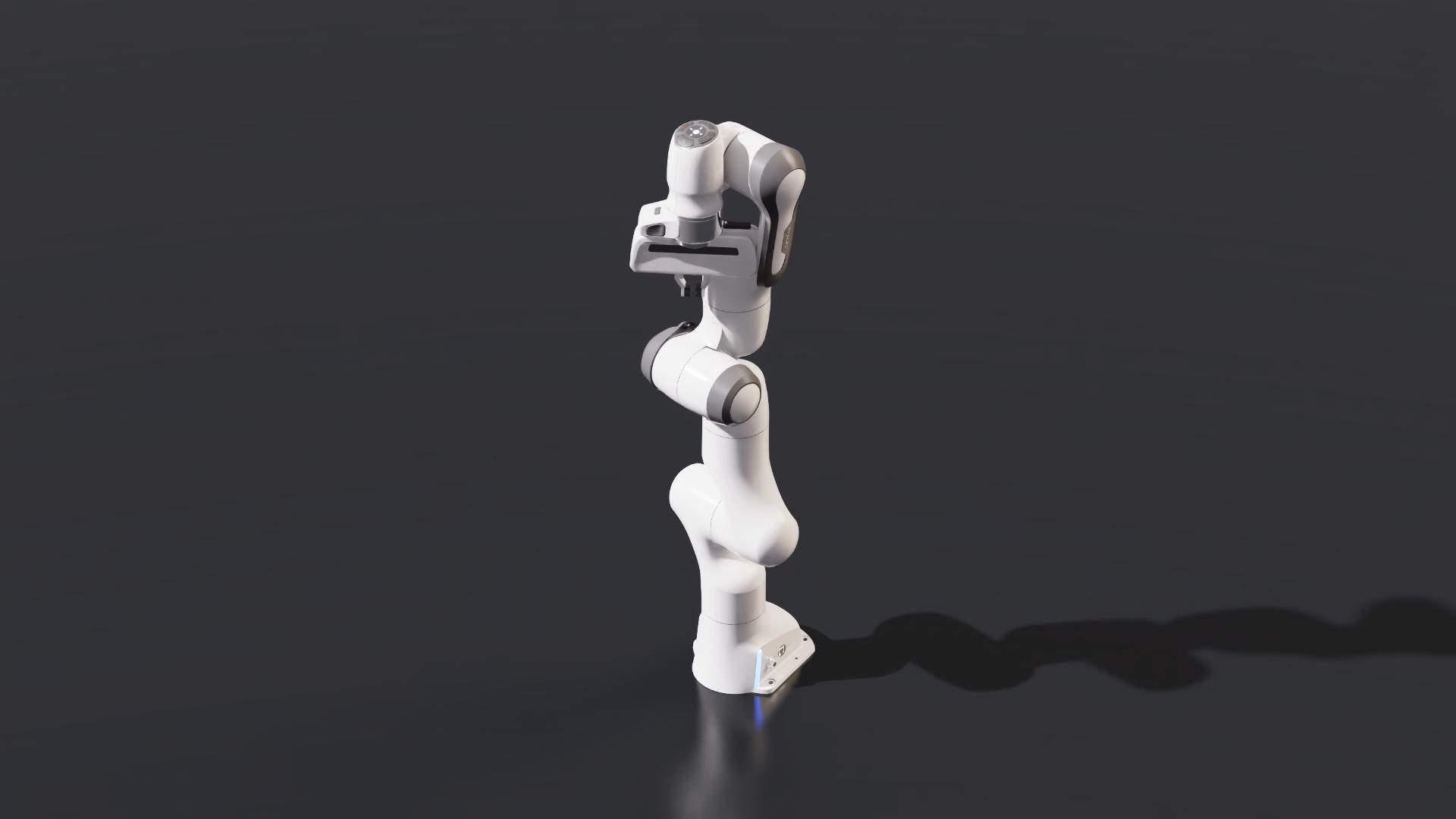}
    \caption{Franka Emika Panda (manipulation).}
    \label{fig:robot_panda}
  \end{subfigure}
  \hfill
  \begin{subfigure}[b]{0.48\linewidth}
    \centering
    \includegraphics[width=\linewidth]{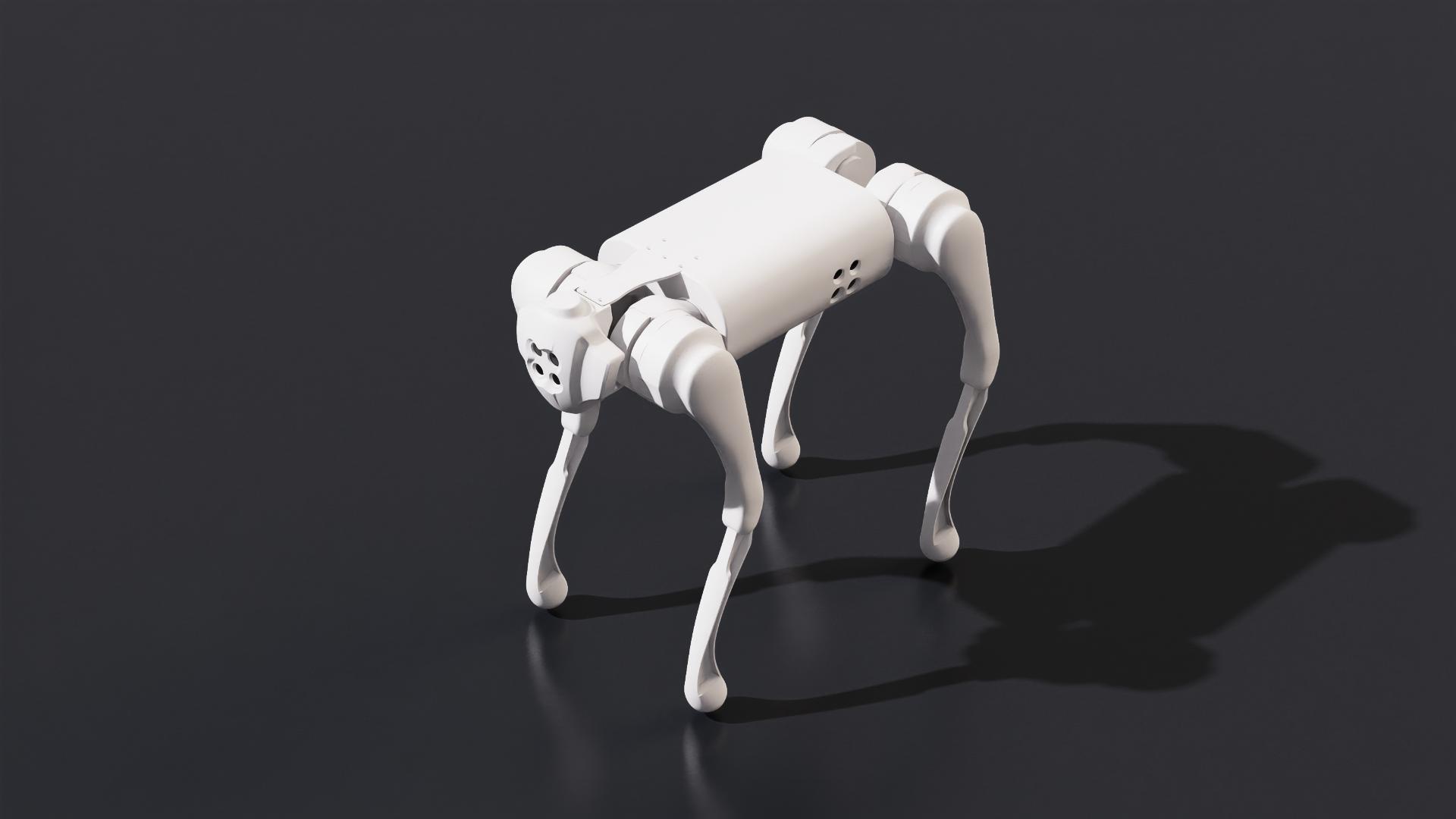}
    \caption{Unitree Go1 (locomotion).}
    \label{fig:robot_unitree}
  \end{subfigure}
  \caption{\textbf{Robot configurations supported in EnvDreamer-generated environments.} EnvDreamer ships with a 7-DoF Franka Emika Panda arm for contact-rich manipulation tasks (left) and a Unitree Go1 quadruped for navigation and locomotion tasks (right). Both robots are spawned directly into procedurally generated EnvDreamer scenes and share a common sensor and action contract, making the dataset suitable for training general-purpose embodied policies.}
  \label{fig:robots}
\end{figure}

EnvDreamer-20k ships with two ready-to-use robot configurations covering manipulation and locomotion (Figure~\ref{fig:robots}), and the framework supports robotics throughout: every generated environment is fully interactive and physics-enabled, exposes the same RGB/depth/proprioception observation contract, and allows policies trained inside EnvDreamer to be deployed on real hardware via image-conditioned reconstruction of the target scene. Both robots share a common observation and action interface so that policies can be reused across embodiments with minimal changes.

\paragraph{Franka Emika Panda (manipulation).}
A 7-DoF Franka arm with a parallel-jaw gripper, used for pick-and-place, tool use, and contact-rich rearrangement tasks. The arm is mounted on a configurable base (fixed table or wheeled platform), and policies issue end-effector pose targets that are resolved by a UE5-side IK solver to per-joint commands. The ArmPointNav and Room Rearrangement experiments in Section~\ref{sec:experiments} use this configuration.

\paragraph{Unitree Go1 (locomotion).}
A quadrupedal locomotion platform used for ObjectNav and exploration tasks. Discrete navigation actions emitted by the policy are mapped to a locomotion controller, and egocentric RGB observations match those produced by the Franka configuration so that a shared visual backbone can drive both embodiments. The four ObjectNav benchmarks reported in Section~\ref{sec:experiments} use this configuration.

\paragraph{Shared sensor and action contract.}
Both robots expose RGB and depth observations, semantic and instance masks, proprioception (joint positions, and end-effector pose where applicable), and the same domain-randomization hooks (textures, lighting, clutter, and selected physics parameters such as friction and mass). New robot platforms can be added by registering a controller and matching this contract through the RC API described in Section~\ref{sec:supp_api}.

\subsection{Training and Evaluation Protocol}
\label{sec:supp_training}

The main paper reports the high-level RL setup. Here we add details that
are mainly of interest for reproduction and extension.

\subsubsection{Compute Budget and Parallelism}

All pretraining and evaluation runs reported in the main paper were
executed on a small cluster of NVIDIA RTX 5090 GPUs using synchronous PPO.
Environments are stepped in parallel across 16--32 workers, with each
worker hosting multiple UE5 instances where hardware permits.
We cap the wall-clock time per update to avoid stragglers; experience
from slow workers in a given round is discarded rather than delaying
gradient updates.

\paragraph{Rendering and throughput.}
EnvDreamer instances are capped at 120 FPS per environment to maintain
stable physics simulation. While a single GPU efficiently runs 3 parallel
instances at full framerate, scaling to multiple GPUs introduces CPU
bottlenecks that reduce per-instance throughput. Despite these limitations,
the system achieves aggregate framerates sufficient for large-scale
training. Table~\ref{tab:rendering_speed} reports measured throughput
across configurations.

\begin{table}[t]
\centering
\footnotesize
\resizebox{\linewidth}{!}{
\begin{tabular}{lcccc}
\toprule
Configuration & Instances & FPS/Instance & Total FPS & Efficiency \\
\midrule
1 GPU (3 inst.) & 3 & 119.8$\pm$2.1 & 359.4$\pm$6.3 & 100\% \\
1 GPU (6 inst.) & 6 & 86.7$\pm$4.2 & 520.2$\pm$25.2 & 72.5\% \\
1 GPU (12 inst.) & 12 & 53.6$\pm$6.8 & 643.2$\pm$81.6 & 44.7\% \\
4 GPUs (48 inst.) & 48 & 40.8$\pm$5.3 & 1{,}958.4$\pm$254.4 & 34.2\% \\
8 GPUs (96 inst.) & 96 & 28.9$\pm$5.7 & 2{,}774.4$\pm$547.2 & 24.2\% \\
\bottomrule
\end{tabular}
}
\caption{\textbf{Rendering performance and scaling.} Per-instance and aggregate throughput on NVIDIA RTX 5090 GPUs (AMD Ryzen 9 7950X CPU). Efficiency is relative to ideal linear scaling. Per-instance FPS drops with parallelism due to CPU bottlenecks, but aggregate throughput continues to grow.}
\label{tab:rendering_speed}
\end{table}

\subsubsection{Curriculum and Task Sampling}

Within a given pretraining run, we sample tasks as follows:

\begin{itemize}
  \item \textbf{Stage~1: Navigation-heavy.}  
        The agent is exposed primarily to goal-reaching tasks with sparse
        rewards to bootstrap basic exploration and map-building skills.
  \item \textbf{Stage~2: Rearrangement and short-horizon manipulation.}  
        Once navigation success stabilizes, we gradually increase the
        fraction of pick--place and toggle tasks.
  \item \textbf{Stage~3: Mixed regimes.}  
        The final phase interleaves all available task types to avoid
        over-specialization.
\end{itemize}

Transition points between stages are defined by simple performance
thresholds (e.g., moving from Stage~1 to Stage~2 once the smoothed
navigation success exceeds a preset target).

\subsubsection{Evaluation Details}

For each benchmark, we use the official evaluation scripts and metrics
released by its authors. In addition to the aggregated success and SPL
numbers reported in the main paper, we log per-skill success (e.g.,
``open-only'' vs.\ ``open-and-pick'' tasks) and per-room-type success.

\subsection{Extended Ablations}
\label{sec:supp_ablations}

This section outlines ablations that did not fit into the main text.
We only summarize the trends here.

\subsubsection{Prompt Strictness}

We vary the strictness of the system prompt along two axes:
(1) whether the model is explicitly reminded to keep unspecified rooms and
items unchanged, and
(2) how emphatically we describe the mutual exclusivity of color and
texture fields.

Preliminary experiments suggest that relaxing these constraints reduces
schema violations but increases the frequency of unintended global edits
(e.g., rooms being resized when only a material change was requested).
Ablation results will be shown in \emph{Table~S2}.

\subsubsection{Validator Components}

\begin{figure}[t]
  \centering
  \includegraphics[width=0.9\linewidth]{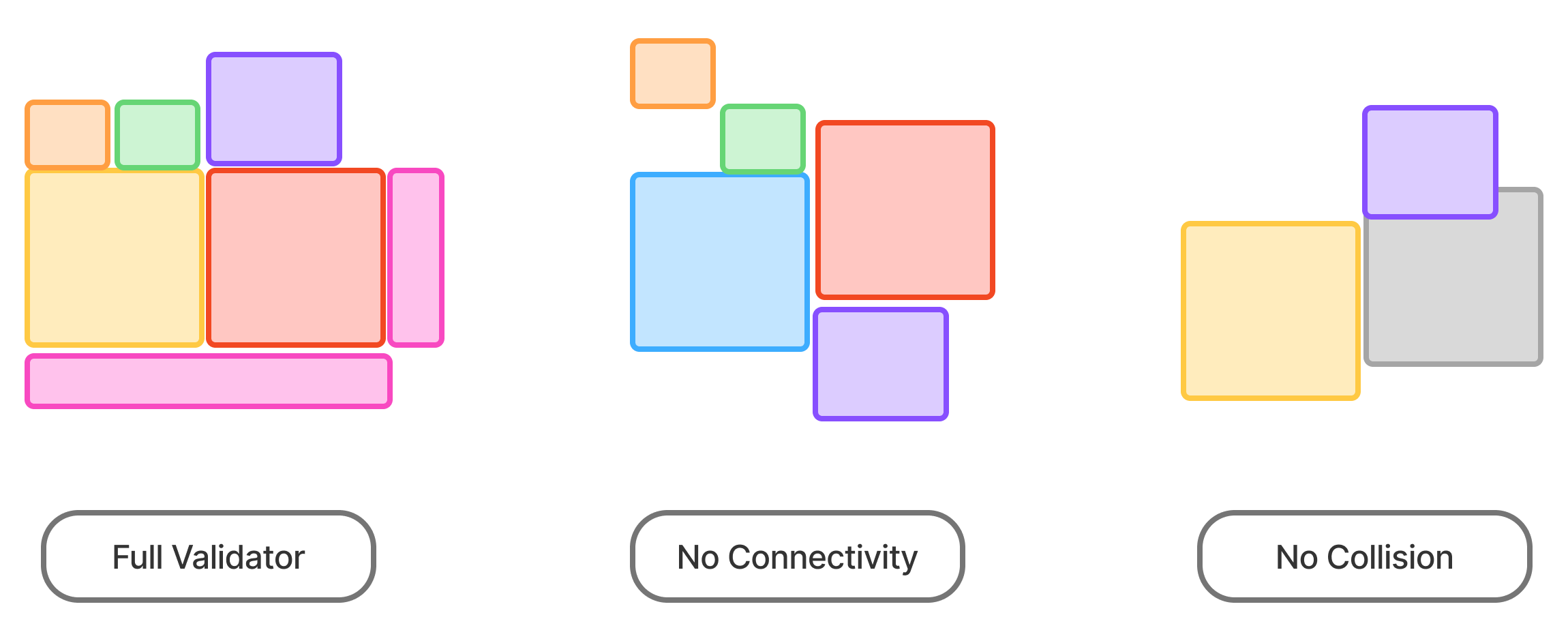}
  \caption{Layouts generated with different subsets of validator checks.}
  \label{fig:validator_ablate}
\end{figure}

We also test the impact of turning off individual validator components:
schema checks, navigation connectivity, and collision checks.
Qualitatively, disabling connectivity checks yields many more ``pretty but
impassable'' layouts, whereas disabling collision checks mainly introduces
localized visual artifacts. Figure~\ref{fig:validator_ablate} is reserved
for side-by-side visualizations.

\subsubsection{Dataset Quality versus Scale}

To isolate the contribution of scene quality from raw dataset size, we
compare policies pretrained on subsets of EnvDreamer--20k generated with
and without the full validator suite.
For each scale, ``raw'' environments skip physics and reachability checks;
``validated'' environments pass all unit-style validators.
Table~\ref{tab:quality_vs_scale} shows that even at half the scale,
validated environments yield better downstream performance than larger
collections of unfiltered scenes, confirming that quality is a more
efficient lever than scale alone.

\begin{table}[t]
\centering
\footnotesize
\setlength{\tabcolsep}{3pt}
\resizebox{\linewidth}{!}{%
\begin{tabular}{llcccc}
\toprule
Scale & Type & Arch SR/SPL & Robo SR/SPL & HM3D SR/SPL & AI2 SR/SPL \\
\midrule
10k & Raw       & 23.1$\pm$0.9 / 14.2$\pm$0.7 & 40.5$\pm$1.4 / 17.8$\pm$0.8 & 7.3$\pm$0.5 / 4.4$\pm$0.3 & 68.9$\pm$1.8 / 56.3$\pm$1.5 \\
10k & Validated & 27.3$\pm$1.0 / 16.9$\pm$0.8 & 47.2$\pm$1.5 / 20.8$\pm$0.9 & 8.4$\pm$0.5 / 5.2$\pm$0.4 & 72.6$\pm$1.9 / 60.9$\pm$1.6 \\
20k & Raw       & 28.7$\pm$1.0 / 17.5$\pm$0.8 & 49.8$\pm$1.5 / 21.5$\pm$0.9 & 8.8$\pm$0.5 / 5.4$\pm$0.4 & 73.4$\pm$1.9 / 61.8$\pm$1.6 \\
20k & Validated & 32.9$\pm$1.2 / 20.3$\pm$0.9 & 56.8$\pm$1.7 / 24.5$\pm$1.1 & 9.7$\pm$0.6 / 5.9$\pm$0.4 & 76.9$\pm$2.0 / 65.2$\pm$1.7 \\
\bottomrule
\end{tabular}%
}
\caption{\textbf{Dataset quality vs.\ scale.}
Policies pretrained on raw (unvalidated) vs.\ validated environments at two
scales.  Validated environments consistently outperform raw ones of the same
scale, and 10k validated environments match or exceed 20k raw ones, showing
that scene quality is a more efficient lever than raw dataset size.}
\label{tab:quality_vs_scale}
\end{table}

\subsubsection{Validator Strictness}

EnvDreamer--20k scenes are generated under unit-style validators that check
physics consistency, reachability, and goal completeness. We study the effect
of varying validator strictness by parameterizing each check with a threshold
and sweeping over three levels: \emph{loose} (thresholds relaxed to pass
$\approx$90\% of candidates), \emph{medium} (our default, $\approx$70\%
pass rate), and \emph{strict} ($\approx$50\% pass rate).

Stricter validation improves downstream ObjectNav success by 2--4 points
across benchmarks but roughly doubles the number of generator calls required
to fill a fixed dataset quota. We adopt the medium setting as a pragmatic
balance: it yields a 70\% acceptance rate that keeps generation costs
reasonable while still filtering the most common failure modes---clipped
furniture, disconnected rooms, and unreachable task targets.

\subsubsection{Material Randomization Strength}

EnvDreamer supports per-load variation in materials and lighting.
We sweep over three randomization levels (none, moderate, aggressive)
during pretraining and observe that moderate variation tends to give the
best cross-domain transfer, while aggressive randomization occasionally
hurts downstream performance on visually consistent benchmarks. Numerical
results can be inserted into \emph{Table~S3}.

\subsection{Qualitative Visualizations and Failure Modes}
\label{sec:supp_qual}

\subsubsection{Additional Layout Samples}

\begin{figure}[t]
  \centering
  \includegraphics[width=0.9\linewidth]{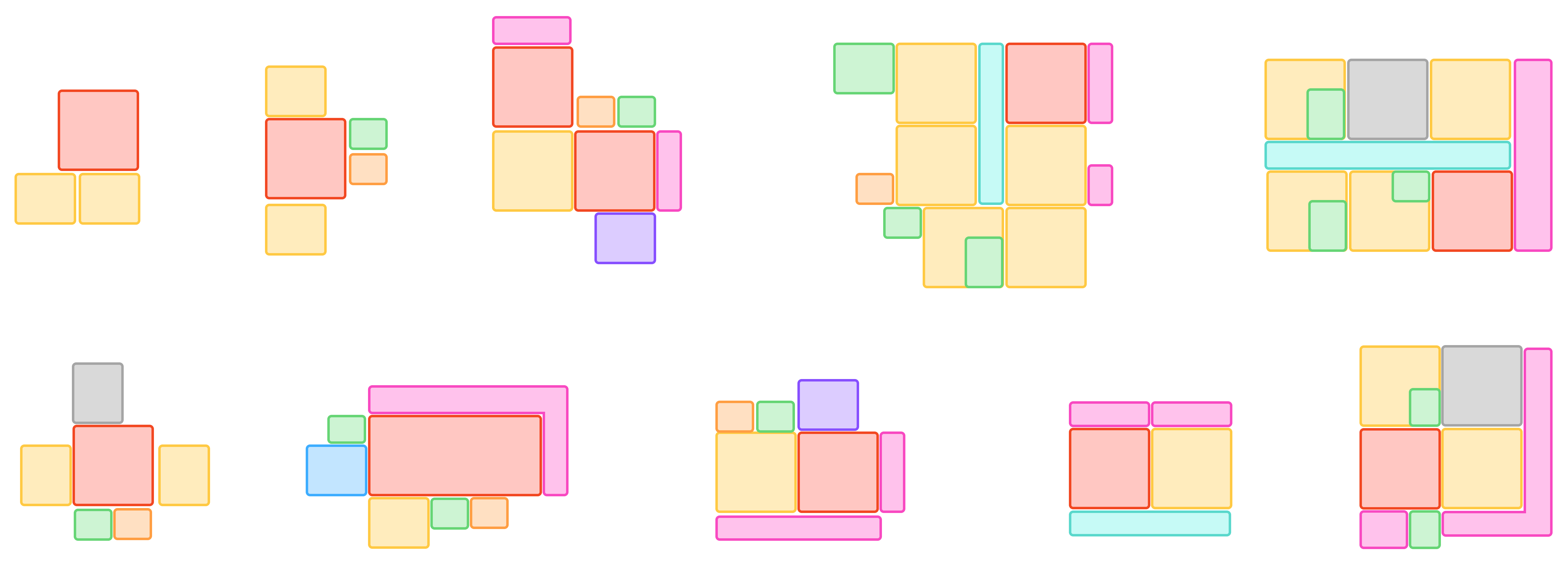}
  \caption{Additional EnvDreamer layouts conditioned on diverse prompts.}
  \label{fig:layout_gallery}
\end{figure}

Figure~\ref{fig:layout_gallery} is for a gallery
of EnvDreamer layouts across a range of prompts. 


\subsubsection{Failure Modes}

Finally, we collect a small set of representative failure cases, grouped
by underlying cause:

\begin{itemize}
  \item \textbf{Under-specified prompts.}  
        Prompts like ``make the apartment look more modern'' can lead to
        large-scale material changes that inadvertently reduce task
        feasibility (e.g., removing surfaces needed for placement).
  \item \textbf{Ambiguous reference images.}  
        When the input photograph contains mirrors or glass walls, the
        planner occasionally hallucinates additional rooms or openings.
  \item \textbf{Validator blind spots.}  
        Rare configurations can pass all automatic checks while still
        being awkward (e.g., a chair technically reachable but visually
        embedded in a corner).
\end{itemize}

\begin{figure}[t]
  \centering
  \includegraphics[width=0.9\linewidth]{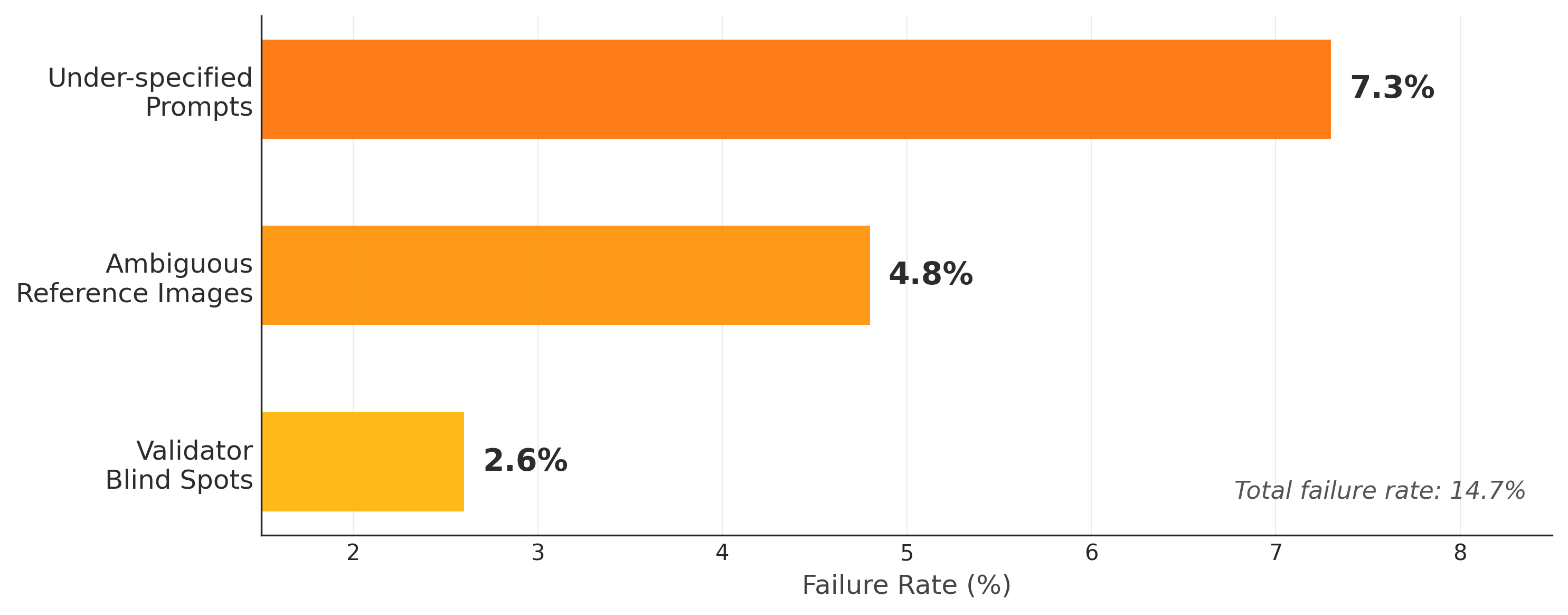}
  \caption{Representative EnvDreamer failure cases grouped by cause.}
  \label{fig:failure_grid}
\end{figure}

\subsection{House Building Engine}
\label{sec:supp_house}

The layout planner produces a JSON description of rooms, bounds, and
materials. The \texttt{HomeAgent} module is responsible for compiling this
description into a concrete UE5 scene by instantiating parameterized
primitives, assigning textures, and inserting door transitions.

\begin{figure}[t]
  \centering
  \includegraphics[width=0.9\linewidth]{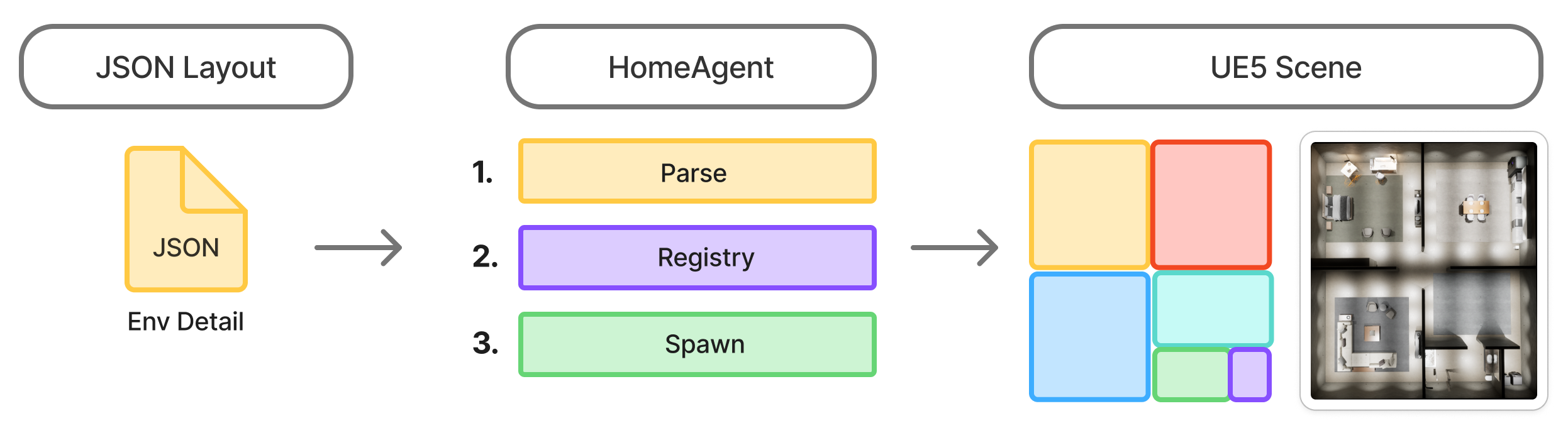}
  \caption{Overview of the house building engine implemented by
  \texttt{HomeAgent}.}
  \label{fig:home_pipeline}
\end{figure}

\subsubsection{Texture Registry and Material Shortcuts}

To keep layouts compact and decoupled from engine paths,
\texttt{HomeAgent} uses a small \emph{texture registry}. At import time,
\texttt{\_load\_texture\_registry()} reads
\texttt{Config/texture\_registry.json} and caches a mapping
from user-friendly shortcuts (e.g., \texttt{"wood\_light"},
\texttt{"concrete"}) to full Unreal material paths:
\begin{itemize}
  \item The registry is loaded once at module level into
        \texttt{\_TEXTURE\_REGISTRY} to avoid repeated I/O.
  \item \texttt{\_get\_texture\_path(name)} resolves a shortcut to a full
        path, or returns the input unchanged if no mapping is found. This
        makes the system robust to both abstract labels and explicit
        engine paths.
\end{itemize}
Textures are applied lazily: after a cube is spawned,
\texttt{apply\_texture(id, texture\_name)} issues an
\texttt{RC\_SetMaterialTexture} call with the resolved material path.
When a texture is present, the agent intentionally omits per-vertex color
parameters so that materials control appearance consistently.

\subsubsection{Primitive-Based Geometry}

The engine builds environments entirely from a single cube asset, scaled
and positioned to realize floors and walls:
\begin{itemize}
  \item \textbf{Tile and pivot.}  
        The cube is parameterized in units of a configurable tile size
        (default 100\,cm). A global pivot mode (\texttt{"base"} or
        \texttt{"center"}) controls whether cubes are anchored at their
        bottom face or at their center; this is handled by
        \texttt{\_cube\_z(height, bottom\_z)}.
  \item \textbf{Spawn API.}  
        \texttt{spawn\_cube(id, x, y, z, sx, sy, sz, color)} wraps a
        single call to \texttt{RC\_AddAsset} with position, scale, and
        orientation. Colors are passed only when no texture is specified,
        mirroring the schema convention that color and texture are
        mutually exclusive.
  \item \textbf{Global parameters.}  
        Scene-wide quantities such as wall thickness, floor thickness,
        door width, and building boundaries are read from the layout file,
        allowing different layouts to share the same builder logic.
\end{itemize}

\subsubsection{Wall Construction and Door Carving}

Walls are constructed by scanning each room and each side of its bounding
box. Let \(\texttt{name}\) be the current room and
\(\texttt{side} \in \{\texttt{north},\texttt{south},\texttt{east},\texttt{west}\}\).

\begin{enumerate}
  \item \textbf{Adjacency detection.}  
        \texttt{\_adjacent\_room\_on\_side(name, side)} searches for a
        neighboring room whose bounds touch the current side along one
        axis and overlap along the orthogonal axis. If no neighbor exists,
        the side is exposed to the building boundary.
  \item \textbf{Ownership and door span.}  
        To avoid double walls, a simple lexical tie-break
        \texttt{\_owns\_shared\_wall(a,b)} chooses which room is
        responsible for a shared wall. If \texttt{name} owns the wall,
        \texttt{\_door\_span(name, other, side)} computes a candidate door
        interval along the shared segment using the configured door width.
  \item \textbf{Segment splitting.}  
        The helper \texttt{spawn\_segments\_for\_side} decomposes the full
        wall segment into:
        \begin{itemize}
          \item solo segments with no neighbor (external walls),
          \item shared segments without doors, and
          \item shared segments with a carved-out door interval.
        \end{itemize}
        Each non-empty segment is realized as a scaled cube with the
        appropriate wall color or texture.
  \item \textbf{Door floor patches.}  
        When a door is carved between two rooms, the agent records two
        small floor patches---one for each room---with their respective
        floor colors or textures. These patches are not spawned
        immediately; instead they are accumulated in
        \texttt{door\_floor\_patches} and instantiated in the floor
        pass, ensuring consistent layering.
\end{enumerate}

\begin{figure}[t]
  \centering
  \includegraphics[width=0.9\linewidth]{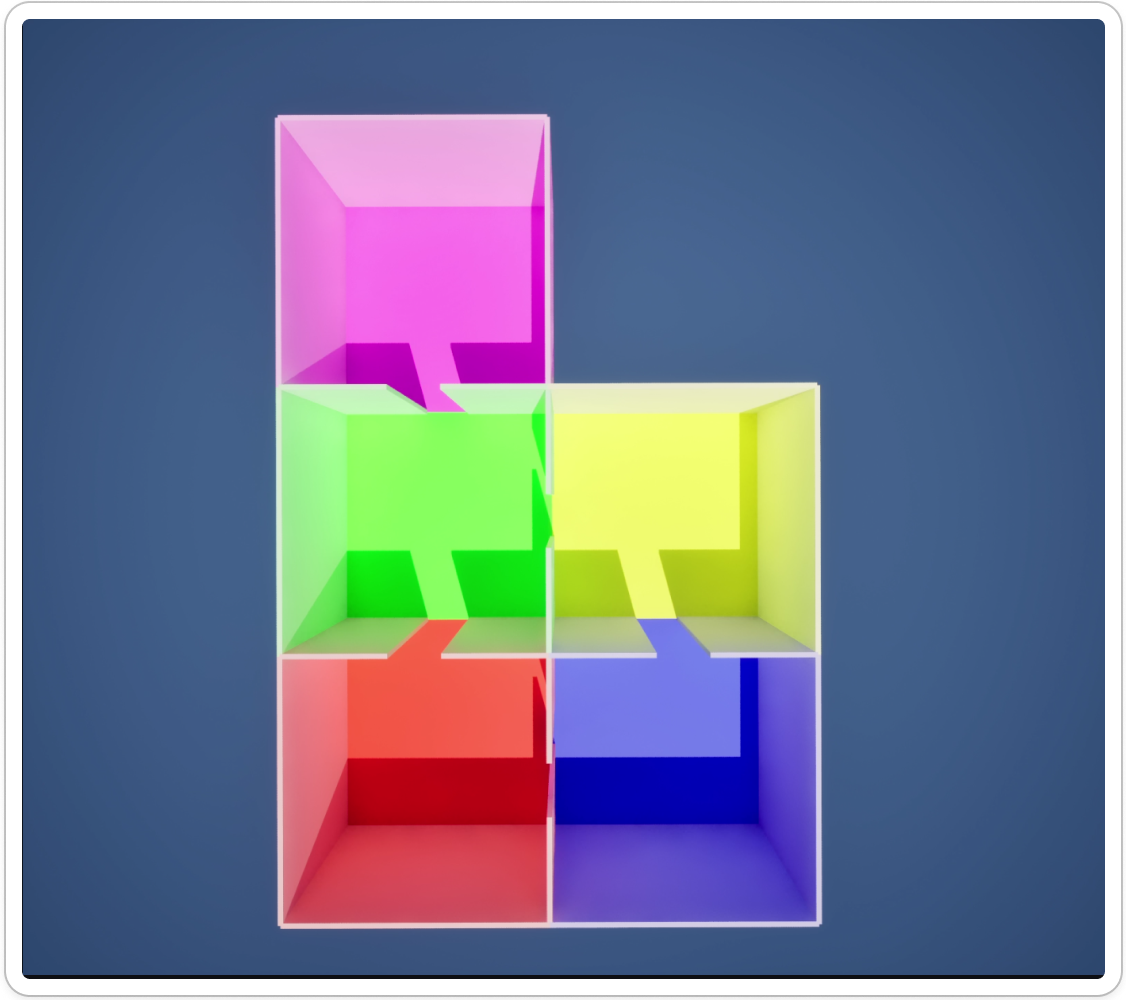}
  \caption{Wall segmentation and door carving, including door floor
  patches.}
  \label{fig:wall_floor_diagram}
\end{figure}

\subsubsection{Floor Slabs and Insets}

Floors are generated in a second pass via \texttt{build\_all\_floors()}:

\begin{itemize}
  \item For each room, the builder computes an inner rectangle obtained by
        insetting the bounds by half the wall thickness. This prevents
        floors from protruding through the walls while keeping a clean
        visual seam.
  \item If the inset would collapse the interior (e.g., extremely thin
        rooms), the code gracefully falls back to using the full bounds.
  \item A single cube is spawned per room to represent its floor slab,
        with either a solid color or a texture, matching the layout’s
        \texttt{floor\_color} / \texttt{floor\_texture} fields.
  \item Door floor patches recorded during wall construction are then
        instantiated as additional thin slabs, using per-room floor
        settings. This produces visually continuous transitions between
        rooms while preserving room-specific appearances.
\end{itemize}

The result is a watertight arrangement of walls and floors that strictly
respects the planner’s bounds and adjacency relationships, while
remaining easy to edit: small changes in the JSON layout (e.g., moving a
wall by one grid cell) directly translate to different cube placements.

\subsubsection{Agent Interface and Entry Points}

The public entry point \texttt{setup\_home()} exposes a simple interface
for downstream tools and scripts:
\begin{itemize}
  \item The \texttt{layout} argument selects between a default layout
        file, a small test layout, or an arbitrary user-provided JSON.
  \item The \texttt{room} argument enables partial builds (e.g., to
        rapidly iterate on a single room without rebuilding the whole
        apartment).
  \item The \texttt{delay} and \texttt{pivot} parameters control spawn
        pacing (useful for debugging) and cube pivot mode, respectively.
\end{itemize}

The \texttt{HomeAgent.clear\_home()} and \texttt{move\_camera()} helpers
are used by interactive front-ends to reset the scene and position the
camera before and after a build, but do not affect the underlying layout.

Overall, this design allows EnvDreamer to compile high-level, LLM-generated
layouts into concrete UE5 scenes using a small, deterministic Python
module, while keeping textures and assets configurable via external
registries.

\newpage
\section*{NeurIPS Paper Checklist}

\begin{enumerate}

\item {\bf Claims}
    \item[] Question: Do the main claims made in the abstract and introduction accurately reflect the paper's contributions and scope?
    \item[] Answer: \answerYes{}
    \item[] Justification: The abstract and Section~1 enumerate three concrete contributions (the EnvDreamer framework, the EnvDreamer-20k dataset, and embodied benchmark evaluation), each of which is supported by Sections~\ref{sec:envdreamer}--\ref{sec:experiments}.

\item {\bf Limitations}
    \item[] Question: Does the paper discuss the limitations of the work performed by the authors?
    \item[] Answer: \answerYes{}
    \item[] Justification: Section~\ref{sec:experiments} is followed by a dedicated ``Discussion and Limitations'' section that covers scope (indoor only), task complexity (short horizons, single solutions), generation latency (2--5 minutes per scene), and CPU-bound throughput when scaling parallel instances.

\item {\bf Theory assumptions and proofs}
    \item[] Question: For each theoretical result, does the paper provide the full set of assumptions and a complete (and correct) proof?
    \item[] Answer: \answerNA{}
    \item[] Justification: The paper presents an environment-generation framework and an empirical dataset evaluation; it makes no theoretical claims that require formal proof.

    \item {\bf Experimental result reproducibility}
    \item[] Question: Does the paper fully disclose all the information needed to reproduce the main experimental results of the paper to the extent that it affects the main claims and/or conclusions of the paper (regardless of whether the code and data are provided or not)?
    \item[] Answer: \answerYes{}
    \item[] Justification: Section~\ref{sec:experiments} specifies benchmarks, splits, evaluation metrics, and the three training regimes used; Supplementary Sections~\ref{sec:supp_training} and~\ref{sec:supp_robots} give compute budget, parallelism, curriculum, and shared sensor/action contracts. Each scene is also stored as a deterministic-seeded program, enabling exact regeneration.

\item {\bf Open access to data and code}
    \item[] Question: Does the paper provide open access to the data and code, with sufficient instructions to faithfully reproduce the main experimental results, as described in supplemental material?
    \item[] Answer: \answerYes{}
    \item[] Justification: We will release the EnvDreamer framework and the EnvDreamer-20k dataset (scene programs, scene graphs, task programs, validator outcomes, and reference trajectories) as stated in Section~\ref{sec:envdreamer100k} and the Conclusion; the Supplementary describes the API surface, web session, and batch materialization scripts needed for reproduction.

\item {\bf Experimental setting/details}
    \item[] Question: Does the paper specify all the training and test details (e.g., data splits, hyperparameters, how they were chosen, type of optimizer) necessary to understand the results?
    \item[] Answer: \answerYes{}
    \item[] Justification: Section~\ref{sec:experiments} states the 16k/2k/2k train/val/test split, the three training regimes, RGB-only inputs, and matched environment-step budgets; Supplementary Section~\ref{sec:supp_training} adds optimizer (synchronous PPO), parallelism (16--32 workers), and curriculum stages.

\item {\bf Experiment statistical significance}
    \item[] Question: Does the paper report error bars suitably and correctly defined or other appropriate information about the statistical significance of the experiments?
    \item[] Answer: \answerYes{}
    \item[] Justification: Tables~\ref{tab:results_procthor_envdreamer},~\ref{tab:scale_ablation_envdreamer}, and~\ref{tab:quality_vs_scale} report mean$\pm$1$\sigma$ values across multiple seeds for each benchmark; Table~\ref{tab:rendering_speed} reports throughput with the same convention.

\item {\bf Experiments compute resources}
    \item[] Question: For each experiment, does the paper provide sufficient information on the computer resources (type of compute workers, memory, time of execution) needed to reproduce the experiments?
    \item[] Answer: \answerYes{}
    \item[] Justification: Section~\ref{sec:experiments} reports that all runs used NVIDIA RTX 5090 GPUs; Table~\ref{tab:rendering_speed} reports per-instance and aggregate throughput across 1--8 GPUs, and Supplementary Section~\ref{sec:supp_training} describes worker counts and wall-clock policies.

\item {\bf Code of ethics}
    \item[] Question: Does the research conducted in the paper conform, in every respect, with the NeurIPS Code of Ethics \url{https://neurips.cc/public/EthicsGuidelines}?
    \item[] Answer: \answerYes{}
    \item[] Justification: The work uses procedurally generated synthetic indoor scenes with no human subjects, no scraped media, and no personally identifiable information; the authors have reviewed the NeurIPS Code of Ethics and the work conforms to it.

\item {\bf Broader impacts}
    \item[] Question: Does the paper discuss both potential positive societal impacts and negative societal impacts of the work performed?
    \item[] Answer: \answerYes{}
    \item[] Justification: Positive impact: scalable simulation environments lower the cost of embodied-AI and robotics research and improve reproducibility. Potential negative impact: better sim-to-real pipelines could accelerate deployment of robotic systems in homes and workplaces; we discuss this and the indoor-only scope as caveats in the Discussion and Limitations section.

\item {\bf Safeguards}
    \item[] Question: Does the paper describe safeguards that have been put in place for responsible release of data or models that have a high risk for misuse (e.g., pre-trained language models, image generators, or scraped datasets)?
    \item[] Answer: \answerNA{}
    \item[] Justification: EnvDreamer-20k consists of synthetic indoor scenes generated from prompts and a curated asset library; it contains no scraped imagery, no personal data, and no pre-trained generative model weights, so it does not pose the misuse risks targeted by this question.

\item {\bf Licenses for existing assets}
    \item[] Question: Are the creators or original owners of assets (e.g., code, data, models), used in the paper, properly credited and are the license and terms of use explicitly mentioned and properly respected?
    \item[] Answer: \answerYes{}
    \item[] Justification: Existing simulators and datasets we compare against (ProcTHOR, AI2-THOR, Habitat, ProcTHOR-10K, RoboTHOR, ArchitecTHOR, HM3D, BEHAVIOR-1K) and tooling (Unreal Engine~5) are credited via citations in Sections~\ref{sec:envdreamer} and the Related Work; license terms for each external asset will be reproduced in the dataset README accompanying the release.

\item {\bf New assets}
    \item[] Question: Are new assets introduced in the paper well documented and is the documentation provided alongside the assets?
    \item[] Answer: \answerYes{}
    \item[] Justification: Section~\ref{sec:envdreamer100k} (``Release format'') lists every artifact shipped with each EnvDreamer-20k entry (prompt, scene program, scene graph, rendering metadata, task programs, validator outcomes, optional reference trajectories) and Supplementary Sections~\ref{sec:supp_api}--\ref{sec:supp_house} document the API, the web session, the planner, and the house-builder pipeline that produced them.

\item {\bf Crowdsourcing and research with human subjects}
    \item[] Question: For crowdsourcing experiments and research with human subjects, does the paper include the full text of instructions given to participants and screenshots, if applicable, as well as details about compensation (if any)?
    \item[] Answer: \answerNA{}
    \item[] Justification: The paper does not involve any crowdsourcing or research with human subjects; all data are synthetically generated by the EnvDreamer pipeline.

\item {\bf Institutional review board (IRB) approvals or equivalent for research with human subjects}
    \item[] Question: Does the paper describe potential risks incurred by study participants, whether such risks were disclosed to the subjects, and whether Institutional Review Board (IRB) approvals (or an equivalent approval/review based on the requirements of your country or institution) were obtained?
    \item[] Answer: \answerNA{}
    \item[] Justification: No human-subjects research was conducted, so IRB review is not applicable.

\item {\bf Declaration of LLM usage}
    \item[] Question: Does the paper describe the usage of LLMs if it is an important, original, or non-standard component of the core methods in this research? Note that if the LLM is used only for writing, editing, or formatting purposes and does \emph{not} impact the core methodology, scientific rigor, or originality of the research, declaration is not required.
    \item[] Answer: \answerYes{}
    \item[] Justification: LLMs/vLLMs are a core component: a GPT-5--class vLLM serves as the planner that translates natural-language prompts into scene programs and task specifications (Sections~\ref{sec:envdreamer},~\ref{sec:envdreamer100k}, and Supplementary Section~\ref{sec:supp_planner}).

\end{enumerate}

\end{document}